\documentclass{article}

\usepackage{preprint,times}

\usepackage[table, dvipsnames]{xcolor}
\usepackage[utf8]{inputenc} %
\usepackage[T1]{fontenc}    %
\usepackage{url}            %
\usepackage{booktabs}       %
\usepackage{amsfonts}       %
\usepackage{nicefrac}       %
\usepackage{amsmath}
\usepackage{amssymb}
\usepackage{amsthm}
\usepackage{float}
\usepackage{microtype}      %
\usepackage{acronym}
\usepackage{multirow}
\usepackage{pifont}
\usepackage{xspace}
\usepackage{graphicx}       %
\usepackage{subcaption}     %
\usepackage{wrapfig}
\usepackage{enumitem}
\usepackage{hyperref}

\newtheorem{proposition}{Proposition}

\newtheorem*{lemma*}{Lemma}
\newtheorem*{proposition*}{Proposition}

\title{From Feed-Forward to Flow:\\
Unifying Reconstruction and Generation\\
Is Easier Than You Think}

\author{%
Haoru Wang\raisebox{0.6ex}{\scriptsize *}\quad Qianfan Shen\raisebox{0.6ex}{\scriptsize *}\quad Kai Ye\quad Wenzheng Chen\quad Baoquan Chen\\[7pt]
\normalfont\small Peking University\\[3pt]
\normalfont\small\raisebox{0.6ex}{\scriptsize *}Equal contribution.}
\hypersetup{hidelinks,
  pdftitle={From Feed-Forward to Flow: Unifying Reconstruction and Generation Is Easier Than You Think},
  pdfauthor={Haoru Wang, Qianfan Shen, Kai Ye, Wenzheng Chen, Baoquan Chen}}

\begin{document}

\maketitle

\begin{abstract}
Reconstruct where the images provide evidence, and generate where they do not: recent success of spatial world models such as Atlas~\citep{worldlabs2026atlas} highlights the value of unifying reconstruction and generation in one model.
Yet the two have long lived in separate paradigms with distinctive failure modes: feed-forward reconstruction averages ambiguity into blur, while conditional generation invents plausible but scene-inconsistent detail.
In this work, we present a \textbf{unified flow-based formulation} for reconstruction and generation, where a shared clean-target predictor performs direct reconstruction at its single-step endpoint and unfolds conditional generation through multi-step flow.
A controlled toy study reveals the mechanism: with a single step, the predictor collapses to the \textit{conditional mean} just like feed-forward methods, favoring consistency over diversity.
With multi-step inference, the fidelity of generated details grows with \emph{context richness}: closer observations reduce ambiguity and yield better-matched details.
We further instantiate the formulation in appearance and geometry 3D tasks.
JiT-LVSM improves perceptual and distributional quality in novel view synthesis, while JUSt3R retains competitive single-step geometry prediction with additional multi-step inference capabilities that reduces veil and flying-pixel artifacts, producing cleaner surface structure with greater test-time compute.
Together, they show that reconstruction and generation can share both a formulation and a backbone, with their behavior governed by denoising configuration---making unification surprisingly simple.
\end{abstract}

\section{Introduction}
\label{sec:intro}

Inferring a 3D scene from images requires a model both to recover what the observations support and to complete what they leave unknown.
Modern systems typically address these two demands through different paradigms.
Feed-forward reconstruction predicts a target view or scene geometry in a single pass~\citep{charatan2024pixelsplat,jin2025lvsm,wang2024dust3r,wang2025vggt}, whereas conditional generation produces a target through iterative denoising~\citep{watson2022novel,chan2023generative,kani2023upfusion}.
Their characteristic failures are markedly different:  direct prediction can blur appearance, erase fine detail, or place geometry between competing surfaces, while generative inference produces sharp and plausible detail that may not faithfully correspond to the observed scene.
Bridging the boundary between these capabilities is becoming central to general-purpose world models.
Atlas~\citep{worldlabs2026atlas}, for example, brings generation and reconstruction into a shared spatial model: preserving what the observations show while imagining scene content beyond them.
This reflects a broader goal for spatial intelligence---to reconstruct where the observations provide sufficient evidence and generate where information is missing.
Reconstruction and generation should therefore not be treated only as separate tasks, but as complementary behaviors that a unified model must support.
This practical convergence raises a fundamental question: how can the goal of unifying reconstruction and generation be expressed in a common formulation that explains their different behaviors and capabilities?
We address this question with a \textbf{unified flow-based formulation} that is surprisingly simple: direct reconstruction emerges at the one-step endpoint of a shared clean-target predictor, while conditional generation unfolds through multi-step flow.
Specifically, we make this connection precise by placing clean-target prediction along a noise-to-data flow~\citep{lipman2022flow,li2025back}.
The predictor estimates a clean target $X_0$ from the observed context $C$ and an intermediate flow state $X_t$, where the path starts from independent noise at $t=0$ and ends at the target at $t=1$.
Reconstruction and generation thus predict the same target, and differ only in how many predictions are composed along the path.
At the starting endpoint, $X_t$ is pure noise and contains no target-specific information beyond the context.
A one-step prediction from this endpoint therefore has the same Bayes-optimal action as a direct feed-forward predictor under the same supervised task loss.
Under squared-error supervision, this common optimum is the conditional mean $\mathbb{E}[X_0\mid C]$.
The optimality of the mean is classical, but its consequence is easy to overlook: when multiple incompatible targets are consistent with the observations, this point estimate averages their competing details and need not resemble any individual solution.
We refer to this mean-seeking effect as \emph{conditional-mean bias}.
Multi-step flow moves beyond this context-only point estimate by conditioning each prediction on an intermediate flow state.
For the ideal predictor, integrating the resulting state-dependent field transports noise toward the conditional target distribution, allowing an individual plausible solution to emerge.
The consequences are concrete in both appearance and geometry.
In novel view synthesis (NVS), averaging multiple compatible textures or boundaries can erase fine detail and produce blur, while such an averaged prediction may still receive a higher PSNR than a sharper generated view.
In geometry prediction, uncertainty between foreground and background surfaces can place a point estimate between their depths, so that points float in the empty space between surfaces as veils or flying pixels, despite a small pointwise error.
Multi-step flow inference can instead commit to a particular texture or surface, improving perceptual or structural quality even when conventional distortion metrics do not improve.
Such commitment, however, is only as reliable as the available context: sparse observations may leave several solutions compatible with the inputs, while richer observations constrain which generated solution corresponds to the scene.

We study the formulation through a controlled mechanism experiment and two real 3D applications.
First, a toy distribution with an analytic conditional mean verifies that feed-forward and one-step predictions recover the same endpoint solution, while multi-step generation from the same model produces individual conditional samples.
Second, JiT-LVSM instantiates the formulation for novel view synthesis on LVSM~\citep{jin2025lvsm}, improving perceptual and distributional quality and showing how richer context brings generated detail into closer agreement with the reference scene.
Third, JUSt3R carries the same formulation into pointmap prediction on DUSt3R~\citep{wang2024dust3r}: its one-step endpoint remains an effective direct predictor, while multi-step flow reduces veil and flying-pixel artifacts, producing cleaner surface structure even when pointwise depth error does not improve.
Both models retain their feed-forward backbones with only a noisy target input, flow-time conditioning, and flow-matching training, allowing a single checkpoint to act as a direct predictor or generator, where inference steps provide a natural axis for test-time scaling.
Together, these experiments connect the endpoint-to-flow formulation to complementary notions of quality across appearance and geometry.

\paragraph{Contributions.}
\textbf{1)} We present a unified flow-based formulation in which one common predictor supports feed-forward reconstruction and conditional generation through different flow configurations.
\textbf{2)} We identify and explain the conditional-mean bias behind feed-forward prediction, prompting a rethink into the long-prevailing yet potentially misleading distortion scores.
\textbf{3)} We instantiate the formulation in NVS and geometry with JiT-LVSM and JUSt3R, demonstrating simplicity, visual and structural benefits of generative inference.

\section{Unifying Reconstruction and Generation through Flow}
\label{sec:revisit}

We revisit feed-forward prediction as the one-step endpoint of conditional flow. The same clean-target interface describes direct reconstruction and multi-step generation; their difference is how predictions are composed along the path. Formal statements, loss-dependent extensions, and proofs are given in Appendix~\ref{app:proof}.

\subsection{Conditional Clean-Target Prediction}
\label{sec:formulation}

Let $C$ denote the observed context and $X_0\sim p_{\mathrm{data}}(\cdot\mid C)$ the clean target, such as a novel-view image or a pointmap. Following flow matching~\citep{lipman2022flow}, we use the noise-to-data path
\begin{equation}
  X_t=(1-t)\epsilon+tX_0,
  \qquad
  \epsilon\sim p_0,
  \qquad
  \epsilon\perp X_0\mid C,
  \label{eq:general-linear-path}
\end{equation}
where the base noise is drawn independently of $(X_0,C)$. Time runs from noise at $t=0$ to data at $t=1$; $X_0$ always denotes the clean target. The model predicts that target directly~\citep{li2025back}, inducing a flow velocity:
\begin{equation}
  \widehat X_0=f_\theta(X_t,t,C),
  \qquad
  v_\theta(x,t,C)=\frac{f_\theta(x,t,C)-x}{1-t},
  \quad t<1.
  \label{eq:clean-flow-interface}
\end{equation}
Flow matching supervises this velocity against $X_0-\epsilon$, equivalently a time-weighted squared error on the clean prediction. Both reconstruction and generation use this prediction interface.

\subsection{From the Feed-Forward Endpoint to Multi-Step Flow}
\label{sec:single-step-collapse}

\paragraph{Single-step collapse.}
A feed-forward predictor receives only context; at $t=0$, a denoiser additionally receives independent noise. Under the same squared-error loss, both therefore have the same population optimum:
\begin{equation}
  f^\star(\epsilon,0,C)=\mathbb E[X_0\mid\epsilon,C]
  =\mathbb E[X_0\mid C].
  \label{eq:endpoint-conditional-mean}
\end{equation}
One Euler step from noise to the data endpoint returns this prediction directly. This is the \emph{conditional-mean bias}: when several targets fit the context, a one-step estimate averages their competing explanations. Other task losses select different summaries, but independent noise still leaves the optimal-action set unchanged (Proposition~\ref{prop:collapse}).

\paragraph{Multi-step flow.}
\label{sec:multi-step-escape}
At intermediate times, the training state contains target information, so the optimal squared-error predictor $\mathbb E[X_0\mid X_t,C]$ depends on the current state. The corresponding ideal velocity transports noise toward the conditional target distribution. Thus a sequence of conditional means need not produce the context-only mean: a single jump returns a summary, while integrating the state-dependent field can generate a sample.
At inference, intermediate states propagate the model's evolving hypothesis; they do not supply new observations. Denoising depth controls how many prediction steps compose that hypothesis. A model trained along the flow can therefore span reconstruction-like and generation-like behavior through its inference configuration.

\subsection{Implications for 3D Prediction}
\label{sec:posterior-width}

When context leaves several plausible explanations, averaging can blur image detail or compromise surface structure. It can nevertheless improve pointwise distortion, so PSNR alone may reward the very behavior that makes a prediction look worse. Our toy experiment makes this mechanism visible, and the NVS and geometry experiments examine its perceptual and structural consequences.

Context and depth play complementary roles. Richer observations constrain which scene explanations are plausible; additional flow steps express the ambiguity that remains. Under sparse context, a sharp sample may still choose an incorrect completion. With richer context, generated detail can agree more closely with the observed scene. This shared relationship applies to both appearance and geometry, and gives multi-pass denoising a practical role as test-time computation.

\section{Unifying Feed-Forward Reconstruction and Generation}
\label{sec:unify}

We test the unified flow formulation in three settings that span a range of posterior widths. A toy distribution with an exact posterior
lets us vary its width directly and compare against analytic references
(\ref{sec:toy}). Novel view synthesis has a broad posterior whose
width varies with source-view overlap. We convert LVSM into JiT-LVSM and compare the two under sparse and rich context (\ref{sec:denoising-depth-exp}).
Pointmap prediction has a narrow posterior that is ambiguous mainly at
depth discontinuities. We convert DUSt3R into JUSt3R and evaluate one
checkpoint with one and multiple steps (\ref{sec:dust3r}). In both 3D
tasks we read distortion metrics together with perceptual or structural ones, then demonstrate how a common prediction interface supports reconstruction and generation across different output spaces.
\subsection{Toy Experiment}
\label{sec:toy}

\begin{figure}[H]
  \centering
  \setlength{\tabcolsep}{1pt}
  \renewcommand{\arraystretch}{0.92}
  \makebox[\linewidth][c]{%
  \resizebox{1.0\linewidth}{!}{%
  \begin{tabular}{c ccccc @{\hspace{4pt}} ccccc}
     & {\tiny\texttt{gt}} & {\tiny\texttt{mean}} & {\tiny\shortstack{\texttt{feed-}\\\texttt{forward}}} & {\tiny\texttt{flow-1}} & {\tiny\texttt{flow-15}}
     & {\tiny\texttt{gt}} & {\tiny\texttt{mean}} & {\tiny\shortstack{\texttt{feed-}\\\texttt{forward}}} & {\tiny\texttt{flow-1}} & {\tiny\texttt{flow-15}} \\
    \raisebox{0.040\linewidth}{\tiny\texttt{r=0.4}} &
    \includegraphics[width=0.085\linewidth]{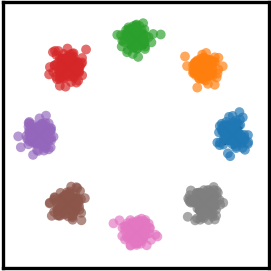} &
    \includegraphics[width=0.085\linewidth]{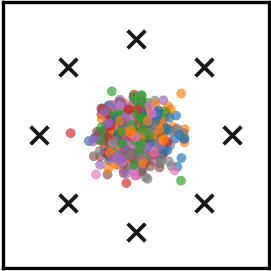} &
    \includegraphics[width=0.085\linewidth]{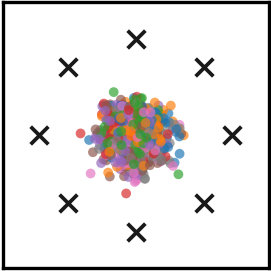} &
    \includegraphics[width=0.085\linewidth]{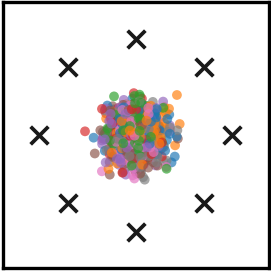} &
    \includegraphics[width=0.085\linewidth]{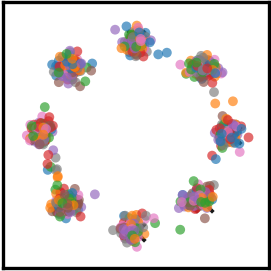} &
    \includegraphics[width=0.085\linewidth]{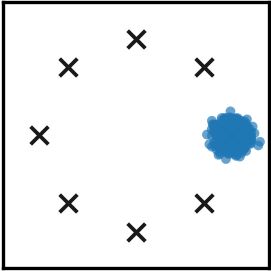} &
    \includegraphics[width=0.085\linewidth]{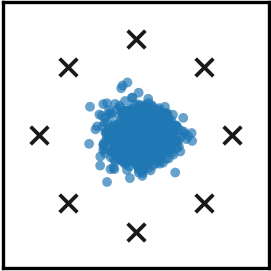} &
    \includegraphics[width=0.085\linewidth]{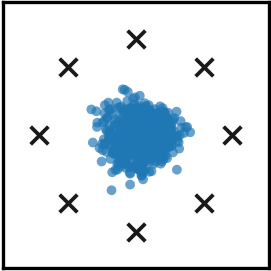} &
    \includegraphics[width=0.085\linewidth]{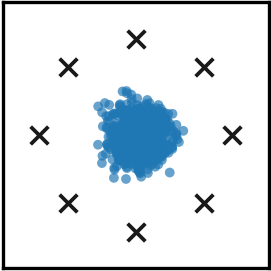} &
    \includegraphics[width=0.085\linewidth]{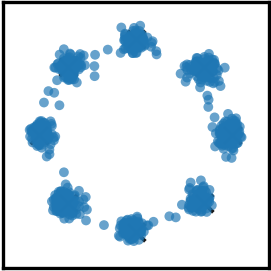} \\
    \raisebox{0.040\linewidth}{\tiny\texttt{r=0.9}} &
    \includegraphics[width=0.085\linewidth]{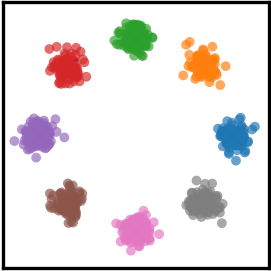} &
    \includegraphics[width=0.085\linewidth]{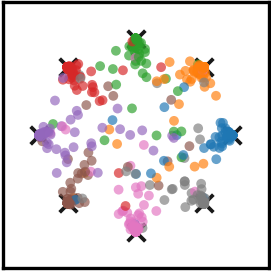} &
    \includegraphics[width=0.085\linewidth]{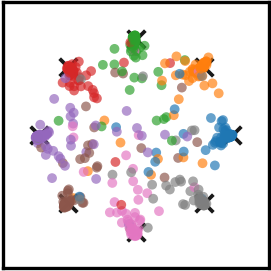} &
    \includegraphics[width=0.085\linewidth]{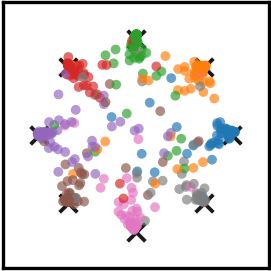} &
    \includegraphics[width=0.085\linewidth]{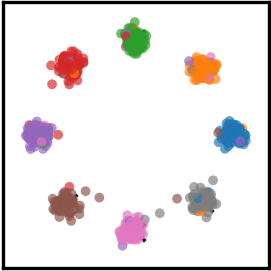} &
    \includegraphics[width=0.085\linewidth]{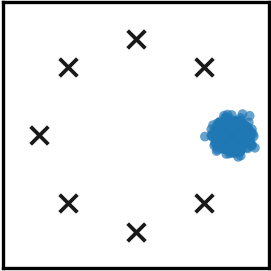} &
    \includegraphics[width=0.085\linewidth]{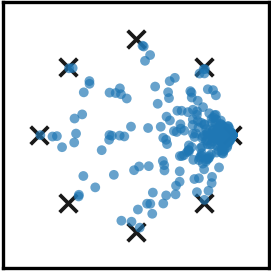} &
    \includegraphics[width=0.085\linewidth]{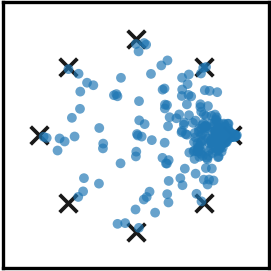} &
    \includegraphics[width=0.085\linewidth]{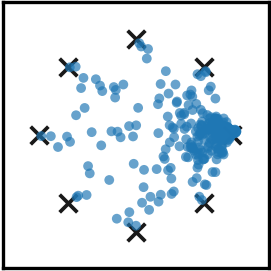} &
    \includegraphics[width=0.085\linewidth]{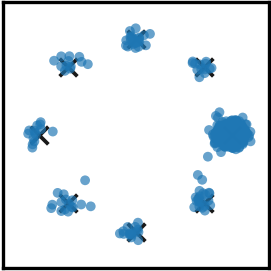} \\
    \raisebox{0.040\linewidth}{\tiny\texttt{r=1.0}} &
    \includegraphics[width=0.085\linewidth]{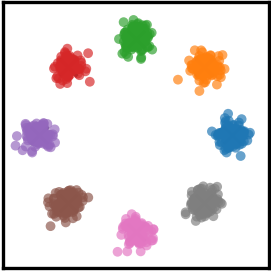} &
    \includegraphics[width=0.085\linewidth]{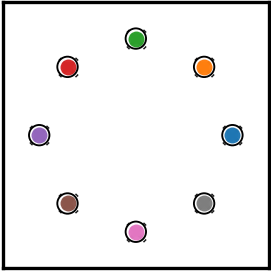} &
    \includegraphics[width=0.085\linewidth]{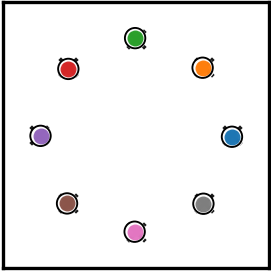} &
    \includegraphics[width=0.085\linewidth]{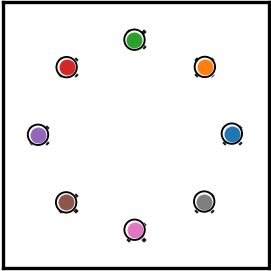} &
    \includegraphics[width=0.085\linewidth]{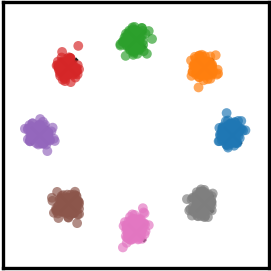} &
    \includegraphics[width=0.085\linewidth]{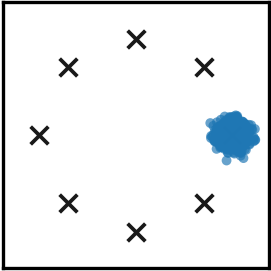} &
    \includegraphics[width=0.085\linewidth]{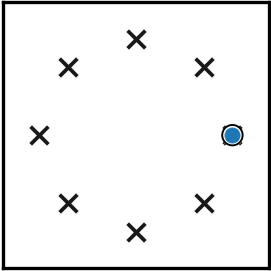} &
    \includegraphics[width=0.085\linewidth]{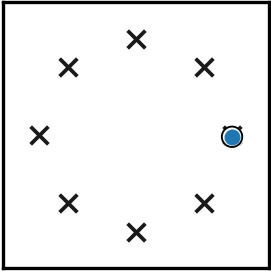} &
    \includegraphics[width=0.085\linewidth]{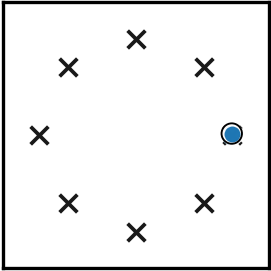} &
    \includegraphics[width=0.085\linewidth]{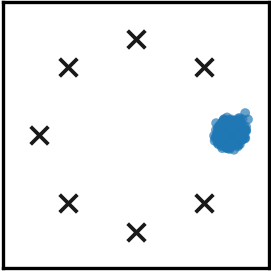}
  \end{tabular}
  }%
  }
  \caption{Controlled toy experiment. Rows increase context reliability $r$. Left blocks show all modes, while right blocks fix one hidden mode for every sample, so wrong-mode samples appear near an incorrect center. Columns: ground truth, the analytic conditional mean, feed-forward prediction, and one flow model evaluated with one and 15 Euler steps. One step tracks the mean, which can lie between modes; 15 steps from the same model sample the posterior.}
  \label{fig:toy}
\end{figure}

\paragraph{Setup.}
We first test the formulation where every quantity of interest is known exactly.
Each sample has a hidden mode $z\in\{0,\dots,K-1\}$ with $K=8$. The target $X_0\in\mathbb{R}^2$ is sampled from an isotropic Gaussian ($\sigma_m=0.15$) centered at~$\mu_z$, where the mode centers~$\mu_k$ lie equally spaced on a circle of radius~2 (marked~$\times$ in Figure~\ref{fig:toy}).
Colors throughout the figure indicate the ground-truth mode~$z$.
A context signal~$c$ noisily identifies the mode, and its reliability $r\in[0,1]$ directly sets the posterior width: at $r=0$ all modes are nearly equally plausible, and at $r=1$ the context identifies the correct one.
Because the mixture is known, both the conditional mean $\mathbb{E}[X_0 \mid c]$ (\texttt{mean}) and exact posterior samples (\texttt{exact}) are available in closed form, providing references for Proposition~\ref{prop:collapse} that are unavailable in the real 3D tasks.

For each~$r$, we train a context-only regressor $f(c)\to X_0$ (\texttt{feed\text{-}forward}) and a single clean-target flow model $f_\theta(X_t,t,c)$ with the same flow-matching objective used in the 3D tasks.
We evaluate the \emph{same} flow checkpoint with one Euler step (\texttt{flow\text{-}1}), which returns $f_\theta(\epsilon,0,c)$, and with 15 Euler steps (\texttt{flow\text{-}15}).
In Figure~\ref{fig:toy}, the left block of each row samples all modes jointly, while the right block fixes a single ground-truth mode for every sample, so that wrong-mode samples are directly visible as points near an incorrect center.

\paragraph{One step recovers the conditional mean.}
In every row of Figure~\ref{fig:toy}, the \texttt{flow-1} column is visually indistinguishable from the \texttt{mean} column, as is \texttt{feed-forward}.
The flow model receives an extra noise input, yet its one-step output collapses onto the same point estimate, as Proposition~\ref{prop:collapse} predicts.

\paragraph{More steps commit to the data manifold.}
At $r=0.4$, the \texttt{mean}, \texttt{feed-forward}, and \texttt{flow-1} columns all place most of their predictions in the interior of the circle, between the modes and far from any ground-truth cluster.
The best squared-error summary thus need not resemble the data it predicts.
Run with 15 steps, the same checkpoint (\texttt{flow-15}) instead places its samples back on the circle, in tight clusters that resemble the \texttt{gt} column.
This is the mechanism behind the metric disagreements in the 3D experiments below: pointwise distortion rewards the averaged prediction, while perceptual and structural measures reveal what it loses.

\paragraph{Posterior width decides whether the commitment is correct.}
The fixed-mode blocks show which modes the samples come from.
At $r=0.4$, \texttt{flow-15} spreads its samples across many modes even though every sample shares one hidden mode: the samples are realistic but often wrong, because the context cannot tell the modes apart.
At $r=0.9$ most samples fall on the correct mode, and at $r=1.0$ all of them do, while the one-step columns converge to the correct center.
As context narrows the posterior, multi-step samples become both realistic and correct.

Quantitative results and experiment details are in Appendix~\ref{app:toy}, further confirming our observations.

\subsection{Novel View Synthesis with JiT-LVSM}
\label{sec:denoising-depth-exp}
\label{sec:lvsm-setup}

\begin{figure}[t]
  \centering
  \includegraphics[width=\linewidth,trim=5 34 5 30,clip]{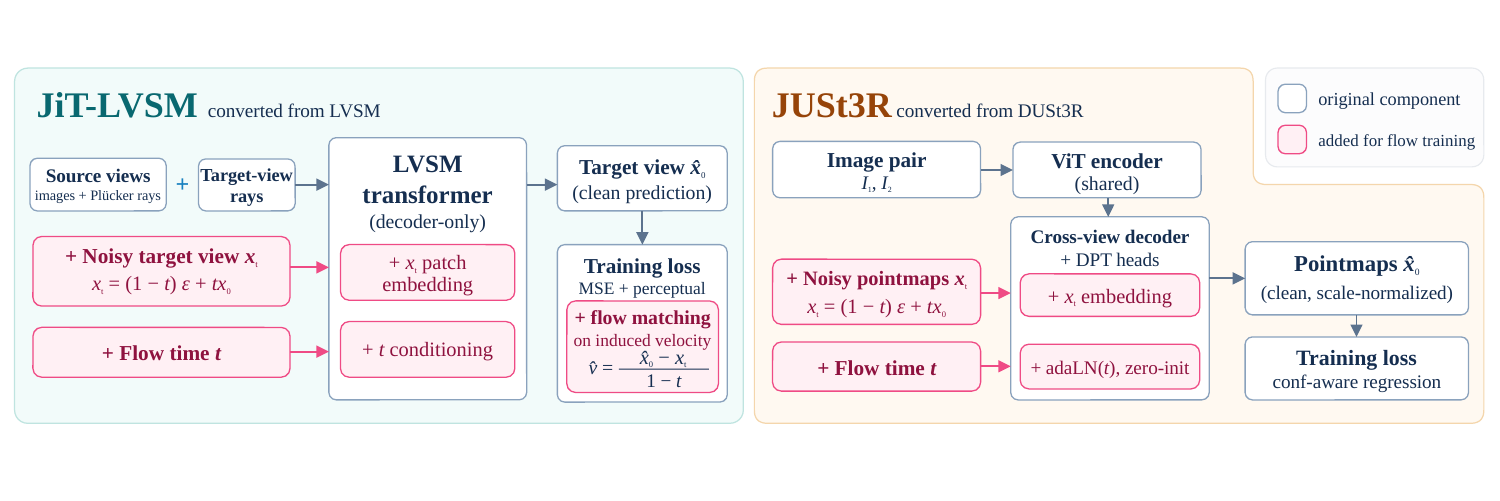}
  \caption{From feed-forward to flow. Both models keep the original backbone and add only a noisy target $x_t$ and flow time $t$; one checkpoint is both a one-step predictor and a multi-step generator.}
  \label{fig:pipeline}
\end{figure}

\begin{table}[b]
  \centering
  \scriptsize
  \setlength{\tabcolsep}{2.5pt}
  \resizebox{\textwidth}{!}{%
  \begin{tabular}{l|cccc|cccc|cccc|cccc}
    \toprule
    \multirow{3}{*}{Method}
    & \multicolumn{8}{c|}{RealEstate10K~\citep{ZhouTFFS18}}
    & \multicolumn{8}{c}{DL3DV~\citep{ling2024dl3dv}} \\
    \cmidrule(lr){2-9}\cmidrule(lr){10-17}
    & \multicolumn{4}{c|}{Sparse Context}
    & \multicolumn{4}{c|}{Rich Context}
    & \multicolumn{4}{c|}{Sparse Context}
    & \multicolumn{4}{c}{Rich Context} \\
    & PSNR $\uparrow$ & SSIM $\uparrow$ & LPIPS $\downarrow$ & FID $\downarrow$
    & PSNR $\uparrow$ & SSIM $\uparrow$ & LPIPS $\downarrow$ & FID $\downarrow$
    & PSNR $\uparrow$ & SSIM $\uparrow$ & LPIPS $\downarrow$ & FID $\downarrow$
    & PSNR $\uparrow$ & SSIM $\uparrow$ & LPIPS $\downarrow$ & FID $\downarrow$ \\
    \midrule
    PixelSplat & 17.18 & 0.566 & 0.496 & 101.2 & 25.51 & 0.867 & 0.126 & 3.102 & \underline{16.78} & \textbf{0.512} & 0.511 & 127.1 & 21.02 & 0.725 & 0.220 & 5.276 \\
    NoPoSplat & 16.02 & 0.511 & 0.588 & 98.23 & 25.46 & 0.854 & 0.137 & 3.017 & 15.17 & \underline{0.497} & 0.602 & 172.3 & 21.50 & \underline{0.751} & 0.237 & 5.145 \\
    \midrule
    LVSM & \textbf{19.78} & \textbf{0.601} & \underline{0.331} & \underline{32.84} & \underline{27.14} & \textbf{0.875} & \underline{0.118} & \underline{2.979} & \textbf{16.81} & 0.410 & \underline{0.482} & \underline{108.7} & \underline{22.29} & 0.727 & \underline{0.201} & \underline{3.996} \\
    JiT-LVSM & \underline{17.94} & \underline{0.593} & \textbf{0.275} & \textbf{21.45} & \textbf{27.77} & \underline{0.872} & \textbf{0.117} & \textbf{2.977} & 15.13 & 0.446 & \textbf{0.421} & \textbf{52.02} & \textbf{22.93} & \textbf{0.769} & \textbf{0.179} & \textbf{3.292} \\
    \bottomrule
  \end{tabular}
  }
  \vspace{2mm}
  \caption{Novel view synthesis on RealEstate10K~\citep{ZhouTFFS18} and DL3DV~\citep{ling2024dl3dv}. PSNR and SSIM measure agreement with the reference view; LPIPS and FID measure perceptual and distributional quality. \textbf{Bold}: best; \underline{underline}: second best.}
  \label{tab:main-lvsm}
\end{table}

\paragraph{Setup.}
We instantiate the formulation with the decoder-only LVSM backbone~\citep{jin2025lvsm} on RealEstate10K~\citep{ZhouTFFS18} and DL3DV~\citep{ling2024dl3dv}.
JiT-LVSM keeps LVSM's source-image and camera-ray interface and adds a noisy target image and flow-time conditioning (Figure~\ref{fig:pipeline}, left); it predicts the clean image and is trained through the induced velocity with flow matching and image-space supervision~\citep{li2025back}.
At inference, it generates the target view by integrating the learned flow from noise with Heun's method in 10 steps.
Both models share the same backbone, resolution, and conditioning inputs, so the comparison isolates the effect of the flow formulation.
We also report PixelSplat~\citep{charatan2024pixelsplat} and NoPoSplat~\citep{ye2024noposplat} as feed-forward references with explicit 3D Gaussian representations.
Architecture and training details, together with the target-side noise sanity check, are given in Appendix~\ref{app:additional}.

\paragraph{Posterior width depends on context.}
When the source views leave the target underconstrained, the conditional target distribution is broad, and it is broad over large regions of the image rather than only at boundaries.
In this regime, LVSM produces conservative, blurred predictions in ambiguous regions, as expected from the averaging mechanism of Section~\ref{sec:single-step-collapse}.
Multi-step flow inference lets JiT-LVSM commit to sharp detail instead, but under sparse context the committed detail may belong to a plausible completion that the observed scene does not support.
Richer context narrows the posterior, so the same inference produces detail that agrees more closely with the reference.
We therefore compare the two models under sparse and rich context.

\definecolor{warmdeeporange}{RGB}{215,95,15}

\begin{figure}[t]
  \vspace{-2mm}
  \centering
  \setlength{\tabcolsep}{0.5pt}
  \renewcommand{\arraystretch}{0.2}
  \newcommand{\imgcell}[1]{\begin{minipage}[t]{0.14\textwidth}\vspace{0pt}\includegraphics[width=\linewidth]{#1}\end{minipage}}
  \newcommand{\inputcell}[1]{\begin{minipage}[t]{\dimexpr0.0675\textwidth\relax}\vspace{0pt}\includegraphics[width=\linewidth]{fig/picks_dl3dv/#1/input_00.png}\\[1pt]\includegraphics[width=\linewidth]{fig/picks_dl3dv/#1/input_01.png}\end{minipage}}
  \newcommand{\pickrow}[3]{
    \inputcell{#1} &
    \imgcell{fig/picks_dl3dv/#1/view#2_baseline.png} &
    \imgcell{fig/picks_dl3dv/#1/view#2_jit.png} &
    \imgcell{fig/picks_dl3dv/#1/view#2_gt.png} &
    \imgcell{fig/picks_dl3dv/#1/view#3_baseline.png} &
    \imgcell{fig/picks_dl3dv/#1/view#3_jit.png} &
    \imgcell{fig/picks_dl3dv/#1/view#3_gt.png}
  }
  \begin{tabular}{c ccc @{\hspace{5pt}} c ccc}
    {\scriptsize Inputs} &
    {\scriptsize LVSM} & {\scriptsize JiT-LVSM} & {\scriptsize Reference} &
    {\scriptsize LVSM} & {\scriptsize JiT-LVSM} & {\scriptsize Reference} \\
    \pickrow{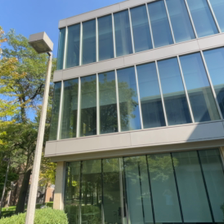}{0083}{0134} \\[1pt]
    \pickrow{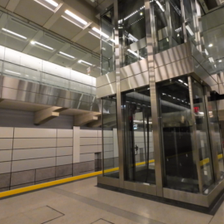}{0069}{0115} \\[1pt]
    \pickrow{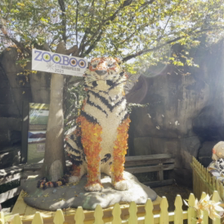}{0049}{0119} \\[1pt]
    \pickrow{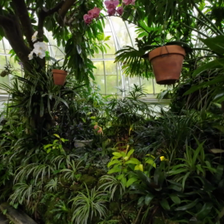}{0047}{0172}
  \end{tabular}
  \caption{Sparse-context comparison on DL3DV~\citep{ling2024dl3dv}. LVSM blurs regions the inputs do not constrain; JiT-LVSM commits to sharp detail. This visual improvement need not raise PSNR, as the aggregate sparse-context results in Table~\ref{tab:main-lvsm} also show.}
  \label{fig:lvsm-sparse}
\end{figure}

\begin{figure}[t]
  \centering
  \setlength{\tabcolsep}{0.5pt}
  \renewcommand{\arraystretch}{0.2}
  \newcommand{\reimgcell}[1]{\begin{minipage}[t]{0.14\textwidth}\vspace{0pt}\includegraphics[width=\linewidth]{#1}\end{minipage}}
  \newcommand{\reinputcell}[1]{\begin{minipage}[t]{\dimexpr0.0675\textwidth\relax}\vspace{0pt}\includegraphics[width=\linewidth]{fig/picks_re10k/#1/input_00.png}\\[1pt]\includegraphics[width=\linewidth]{fig/picks_re10k/#1/input_01.png}\end{minipage}}
  \newcommand{\rerow}[3]{
    \reinputcell{#1} &
    \reimgcell{fig/picks_re10k/#1/view#2_baseline.png} &
    \reimgcell{fig/picks_re10k/#1/view#2_jit.png} &
    \reimgcell{fig/picks_re10k/#1/view#2_gt.png} &
    \reimgcell{fig/picks_re10k/#1/view#3_baseline.png} &
    \reimgcell{fig/picks_re10k/#1/view#3_jit.png} &
    \reimgcell{fig/picks_re10k/#1/view#3_gt.png}
  }
  \begin{tabular}{c ccc @{\hspace{5pt}} c ccc}
    {\scriptsize Inputs} &
    {\scriptsize LVSM} & {\scriptsize JiT-LVSM} & {\scriptsize Reference} &
    {\scriptsize LVSM} & {\scriptsize JiT-LVSM} & {\scriptsize Reference} \\
    \rerow{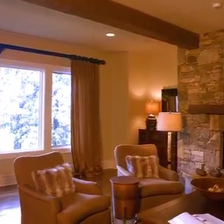}{0101}{0094} \\[1pt]
    \rerow{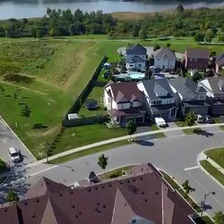}{0276}{0144}
  \end{tabular}
  \caption{Rich-context comparison on RealEstate10K~\citep{ZhouTFFS18}. With more constraining observations, JiT-LVSM's detail agrees more closely with the reference.}
  \label{fig:lvsm-rich}
  \vspace{-4mm}
\end{figure}

\paragraph{Qualitative comparison.}
Figures~\ref{fig:lvsm-sparse} and~\ref{fig:lvsm-rich} show representative scenes.
Across examples, JiT-LVSM produces sharper and more structured target views, while LVSM blurs extrapolated or ambiguous content.
Under sparse context, this blur is a conservative choice. The prediction looks less realistic, but it avoids committing to unsupported high-frequency detail.
JiT-LVSM is more realistic in the same setting, yet some of its added detail does not match the reference view.
Under rich context, the source views constrain the target more tightly, and JiT-LVSM's detail more often agrees with the reference.
Appendix~\ref{app:more-visuals} shows $60$ randomly selected DL3DV scenes and $60$ RealEstate10K scenes.

\paragraph{Quantitative comparison.}
Table~\ref{tab:main-lvsm} compares JiT-LVSM with LVSM~\citep{jin2025lvsm}, PixelSplat~\citep{charatan2024pixelsplat}, and NoPoSplat~\citep{ye2024noposplat} under sparse and rich context on both datasets.
PSNR and SSIM measure pointwise agreement with the reference, which an averaged prediction can score well on. LPIPS and FID measure whether the generated detail looks realistic.
\emph{Sparse context.}
JiT-LVSM gives the best LPIPS and FID on both datasets, but its PSNR is 1.7--1.8\,dB lower than LVSM.
This is the trade-off predicted by Section~\ref{sec:single-step-collapse}: LVSM approximates the PSNR-optimal conditional mean, and a sample from a broad posterior has higher squared error than that mean even when it looks more realistic.
\emph{Rich context.}
The trade-off largely disappears: JiT-LVSM improves PSNR over LVSM by about 0.6\,dB on both datasets and remains best in LPIPS and FID, by a clear margin on DL3DV and a small one on RealEstate10K.
As the posterior narrows, flow inference keeps its realism while its committed detail increasingly agrees with the particular target scene.
\vspace{-2mm}

\paragraph{Measuring context richness.}
\begin{wraptable}{r}{0.45\linewidth}
  \vspace{-4mm}
  \centering
  \scriptsize
  \setlength{\tabcolsep}{3pt}
  \begin{tabular}{lcc}
    \toprule
    Metric & DL3DV & RealEstate10K \\
    \midrule
    \# scenes & $140$ & $6424$ \\
    RoMA mean & $0.111$ & $0.249$ \\
    RoMA median & $0.056$ & $0.237$ \\
    Conf. $>0.5$ mean & $0.049$ & $0.142$ \\
    Grid cov. $>0.5$ med. & $0.188$ & $0.688$ \\
    \bottomrule
  \end{tabular}
  \caption{Source-view overlap measured with RoMA on the two evaluation sets.}
  \label{tab:roma-context}
  \vspace{-4mm}
\end{wraptable}
To make ``sparse'' and ``rich'' measurable, we estimate how much the source views overlap using dense correspondences from RoMA~\citep{edstedt2024roma}, as a proxy for how strongly the inputs constrain the target.
The probe describes the available evidence instead of estimating the posterior entropy directly.
Table~\ref{tab:roma-context} shows that source-view overlap is much weaker in our DL3DV evaluation set than in RealEstate10K: the median fraction of grid cells with confident matches is $0.188$ versus $0.688$.
Weaker overlap indicates a broader posterior, and it is on DL3DV that flow inference changes the output distribution most. FID is roughly halved under sparse context, against a one-third reduction on RealEstate10K.
The next experiment turns to geometry, where the posterior is narrow except at boundaries and the effect of committing to one solution can be measured directly.

\subsection{Geometry Prediction with JUSt3R}
\label{sec:dust3r}

\paragraph{Setup.}
We instantiate the same formulation in pointmap space with DUSt3R~\citep{wang2024dust3r}.
Our \textbf{JUSt3R} keeps DUSt3R's backbone, training data, and schedule, and adds a noisy target pointmap $x_t$ through a light decoder-token embedding and flow time $t$ through adaLN (Figure~\ref{fig:pipeline}, right). It predicts the clean pointmap and is trained with DUSt3R’s own confidence-weighted regression loss.
We denote inference with $k$ steps from a single checkpoint by $\mathrm{JUSt3R}_k$: $k=1$ is the direct-prediction endpoint, and larger $k$ composes predictions along the flow, reusing one context encoding.
We compare against DUSt3R retrained by ourselves, which is the controlled baseline, and report the released checkpoint (DUSt3R$^{\ast}$) for reference.
Implementation and evaluation details are in Appendix~\ref{app:dust3r}.

\newcommand{\ahd}[1]{{\fontsize{5.0}{5.6}\selectfont #1}}
\newcommand{\refc}[1]{\textcolor{gray}{#1}}
\begin{table}[b]
  \centering
  \scriptsize
  \setlength{\tabcolsep}{1.6pt}
  \resizebox{\textwidth}{!}{%
  \begin{tabular}{l|cccc|cccc||cccc|cccc|cccc}
    \toprule
    & \multicolumn{4}{c|}{AbsRel $\downarrow$}
    & \multicolumn{4}{c||}{$\delta_{1.25}$ $\uparrow$}
    & \multicolumn{4}{c|}{BF1 $\uparrow$ ($\times100$)}
    & \multicolumn{4}{c|}{veil $\downarrow$}
    & \multicolumn{4}{c}{flying $\downarrow$} \\
    Benchmark
      & \ahd{\refc{DUSt3R$^{\ast}$}} & \ahd{DUSt3R} & \ahd{JUSt3R$_{1}$} & \ahd{JUSt3R$_{10}$}
      & \ahd{\refc{DUSt3R$^{\ast}$}} & \ahd{DUSt3R} & \ahd{JUSt3R$_{1}$} & \ahd{JUSt3R$_{10}$}
      & \ahd{\refc{DUSt3R$^{\ast}$}} & \ahd{DUSt3R} & \ahd{JUSt3R$_{1}$} & \ahd{JUSt3R$_{10}$}
      & \ahd{\refc{DUSt3R$^{\ast}$}} & \ahd{DUSt3R} & \ahd{JUSt3R$_{1}$} & \ahd{JUSt3R$_{10}$}
      & \ahd{\refc{DUSt3R$^{\ast}$}} & \ahd{DUSt3R} & \ahd{JUSt3R$_{1}$} & \ahd{JUSt3R$_{10}$} \\
    \midrule
    NYU-v2 & \refc{.080} & .075 & \textbf{.070} & \underline{.071} & \refc{.907} & .915 & \textbf{.925} & \textbf{.925} & \refc{5.17} & 7.10 & \underline{8.03} & \textbf{8.84} & \refc{.220} & \underline{.215} & .219 & \textbf{.186} & \refc{.041} & .047 & \underline{.037} & \textbf{.023} \\
    TUM    & \refc{.180} & \underline{.177} & \textbf{.165} & .178 & \refc{.774} & \underline{.750} & \textbf{.782} & .749 & \refc{2.29} & 2.09 & \underline{3.34} & \textbf{4.16} & \refc{.271} & .273 & \underline{.231} & \textbf{.169} & \refc{.027} & .040 & \underline{.038} & \textbf{.022} \\
    Bonn   & \refc{.141} & \textbf{.095} & \underline{.096} & .105 & \refc{.825} & \textbf{.904} & \underline{.902} & .885 & \refc{1.88} & \underline{2.71} & \textbf{2.83} & 2.57 & \refc{.275} & .230 & \underline{.220} & \textbf{.144} & \refc{.016} & .024 & \underline{.021} & \textbf{.014} \\
    Sintel & \refc{.413} & .418 & \textbf{.400} & \underline{.402} & \refc{.560} & .533 & \textbf{.560} & \underline{.558} & \refc{.36} & .16 & \underline{.36} & \textbf{1.24} & \refc{.191} & .210 & \underline{.188} & \textbf{.185} & \refc{.023} & \underline{.030} & \underline{.030} & \textbf{.017} \\
    KITTI  & \refc{.112} & \underline{.098} & \textbf{.096} & .109 & \refc{.863} & \underline{.897} & \textbf{.899} & .873 & \multicolumn{4}{c|}{\emph{sparse GT}} & \multicolumn{4}{c|}{\emph{sparse GT}} & \refc{.109} & .107 & \underline{.106} & \textbf{.089} \\
    \bottomrule
  \end{tabular}}
  \vspace{1mm}
  \caption{Zero-shot depth on NYU-v2~\citep{silberman2012indoor}, TUM~\citep{sturm2012benchmark}, Bonn~\citep{palazzolo2019refusion}, Sintel~\citep{butler2012naturalistic}, and KITTI~\citep{geiger2013vision}. DUSt3R is retrained with our pipeline and is the controlled baseline; JUSt3R$_1$ and JUSt3R$_{10}$ are the same checkpoint with one and ten steps. \textbf{Bold}: best; \underline{underline}: second best, ranked among these three. The released DUSt3R 512-DPT checkpoint (DUSt3R$^{\ast}$, gray) is shown for reference. Veil rate and BF1 require dense ground truth and are omitted on KITTI.}
  \label{tab:dust3r}
\end{table}

\paragraph{Where the geometric posterior is ambiguous.}
Unlike novel view synthesis, pointmap prediction has a posterior that is narrow over most of the image because visible surfaces are well constrained by the input and by learned geometric priors.
Ambiguity concentrates at depth discontinuities, mixed pixels, and thin structures, where a pixel may belong to either the foreground or the background surface.
DUSt3R's loss is a confidence-weighted, scale-normalized $L_{2,1}$ regression, so it still yields a single point estimate per pixel. Near a boundary, where the pixel may belong to either surface and the edge position itself is uncertain, this estimate is often placed between the two surfaces, producing the veils and flying pixels typical of feed-forward pointmap regression.
We therefore expect multi-step inference to change pointwise accuracy little but to act locally at boundaries.

\paragraph{Evaluating surface structure.}
AbsRel and $\delta_{1.25}$ compare each depth against a reference~\citep{wang2025continuous,zhang2025monst3r} and are dominated by the well-constrained interior, so they can hide exactly these boundary errors.
We therefore also report the \emph{veil rate}, the fraction of ground-truth depth discontinuities where the prediction falls in the middle of the gap between the two surfaces. Meanwhile, a ground-truth-free \emph{flying-pixel rate} that flags points on locally grazing geometry~\citep{pomerleau2013comparing} and \emph{BF1} for depth-edge localization~\citep{bochkovskiy2025depth,xu2026pointdit} are included.
Veil and flying-pixel rates reward committing to a surface but not choosing the correct one, so we read them together with BF1 and AbsRel.

\begin{figure}[t]
  \centering
  \setlength{\tabcolsep}{1pt}
  \begin{tabular}{@{}cccc@{}}
    {\scriptsize DUSt3R} & {\scriptsize JUSt3R$_1$} & {\scriptsize JUSt3R$_{10}$} & {\scriptsize JUSt3R$_{50}$} \\[2pt]
    \includegraphics[width=0.244\textwidth]{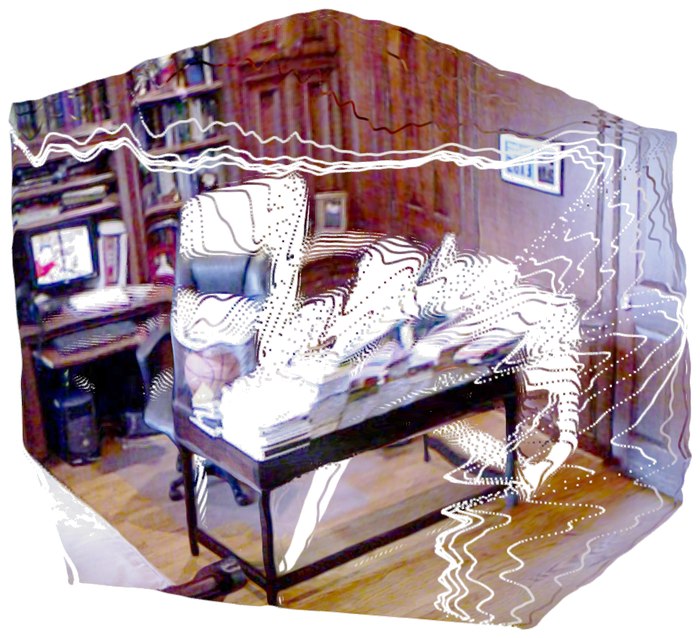} &
    \includegraphics[width=0.244\textwidth]{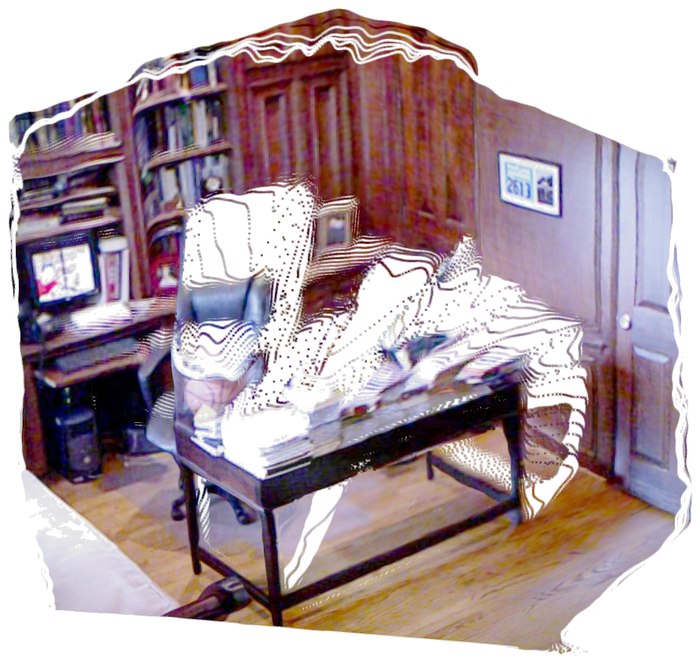} &
    \includegraphics[width=0.244\textwidth]{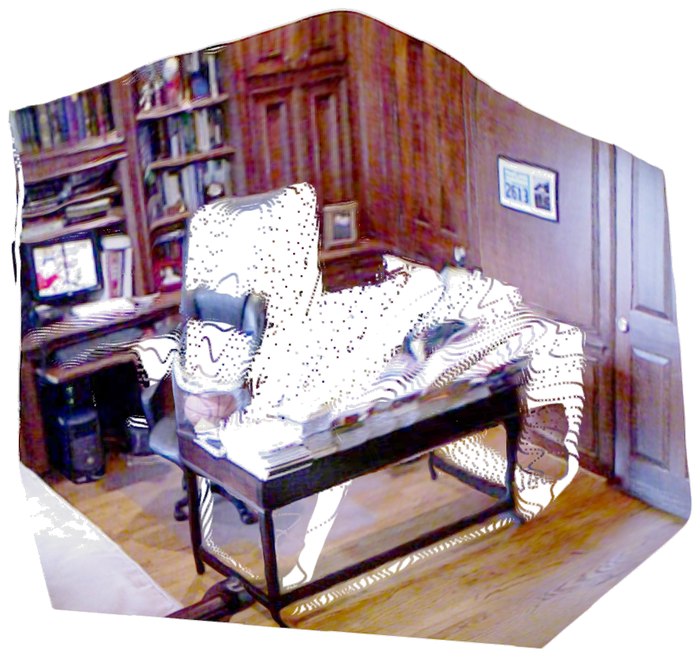} &
    \includegraphics[width=0.244\textwidth]{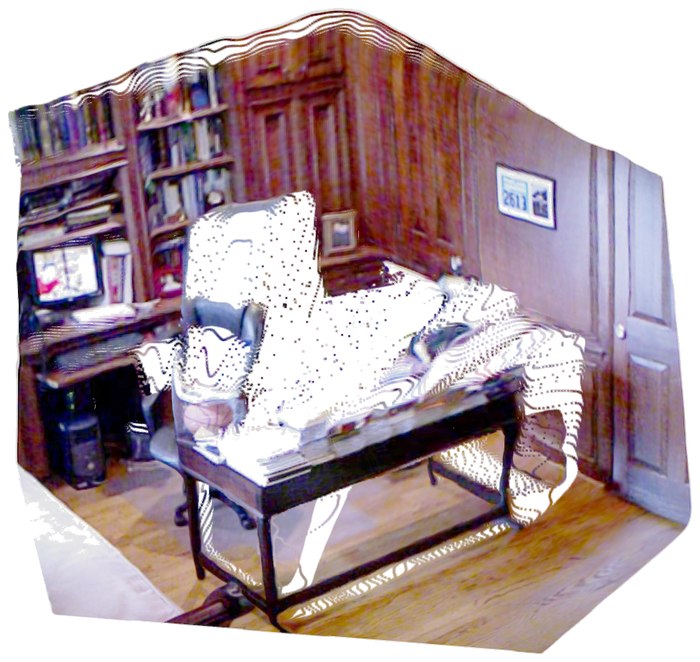} \\[2pt]
    \includegraphics[width=0.244\textwidth]{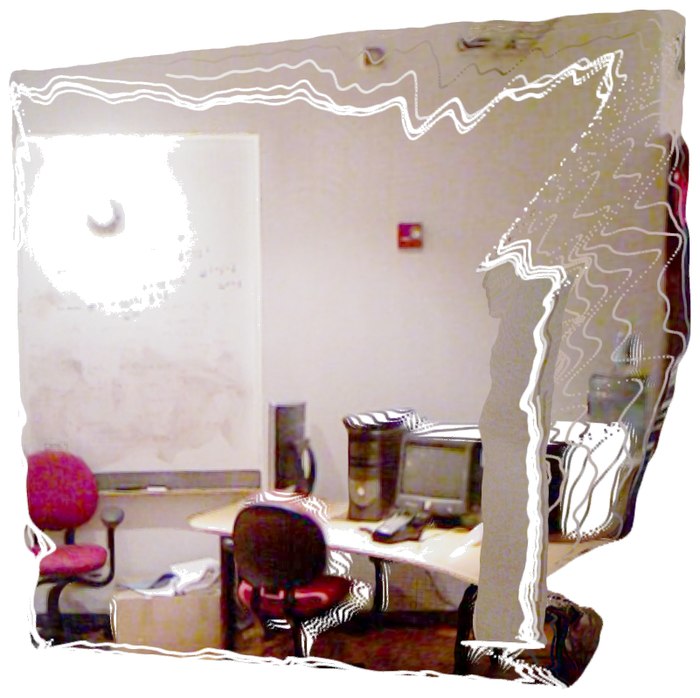} &
    \includegraphics[width=0.244\textwidth]{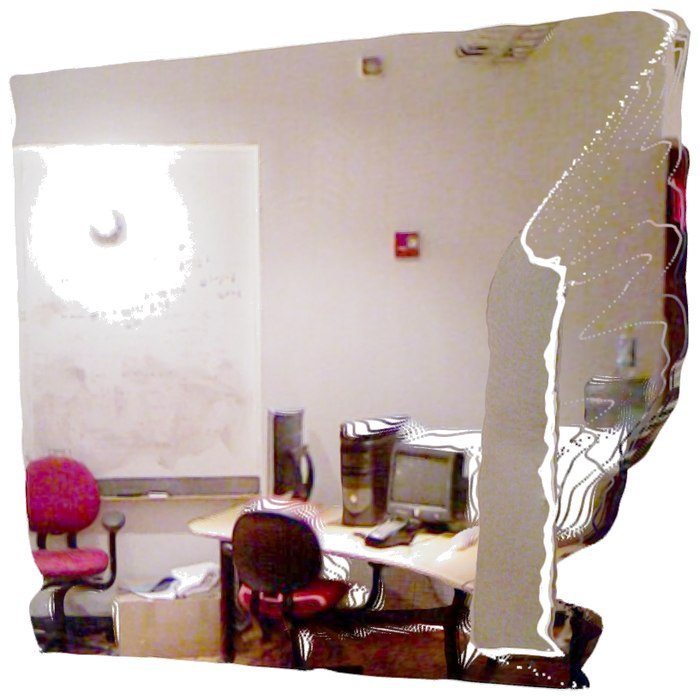} &
    \includegraphics[width=0.244\textwidth]{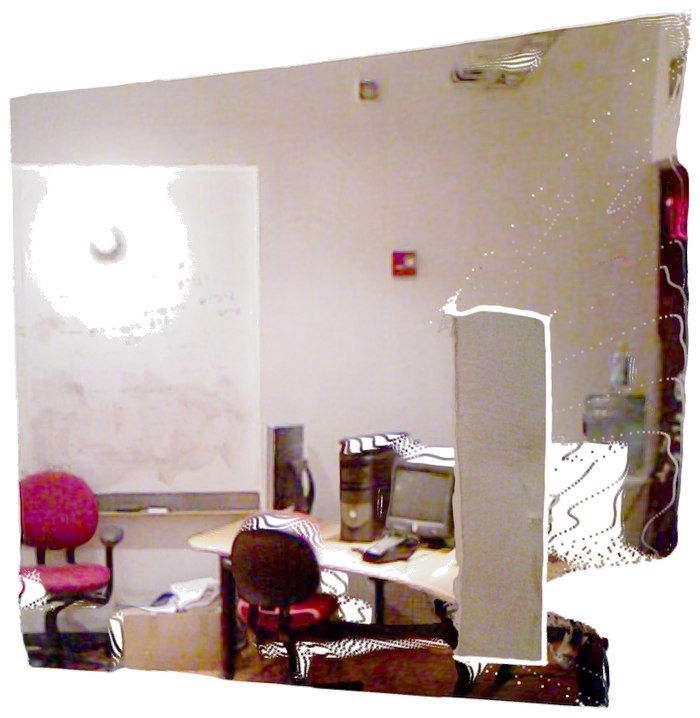} &
    \includegraphics[width=0.244\textwidth]{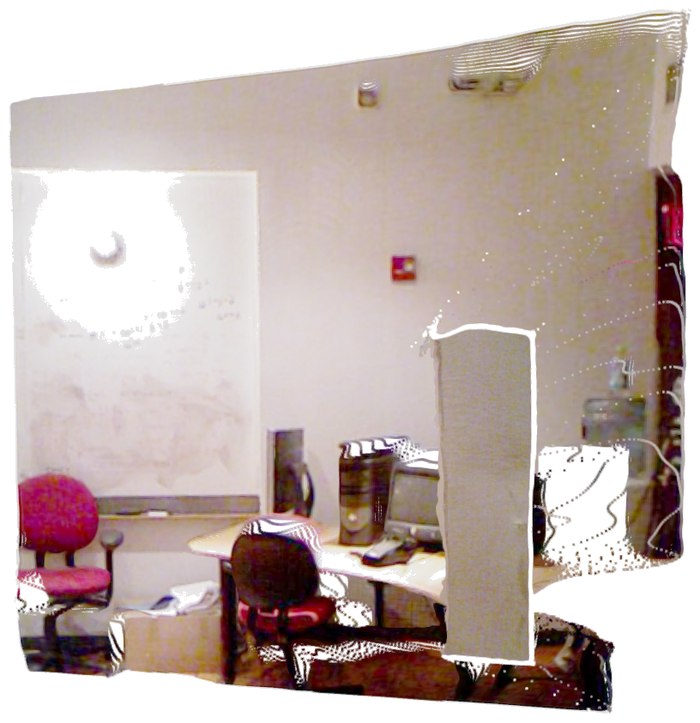} \\[2pt]
    \includegraphics[width=0.244\textwidth]{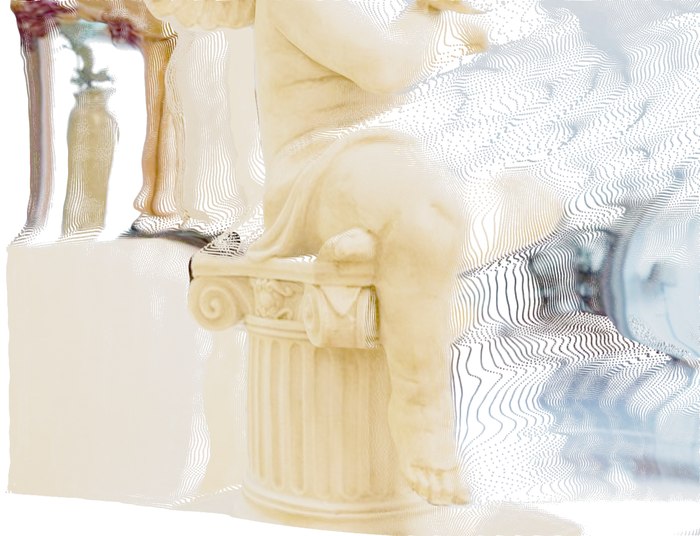} &
    \includegraphics[width=0.244\textwidth]{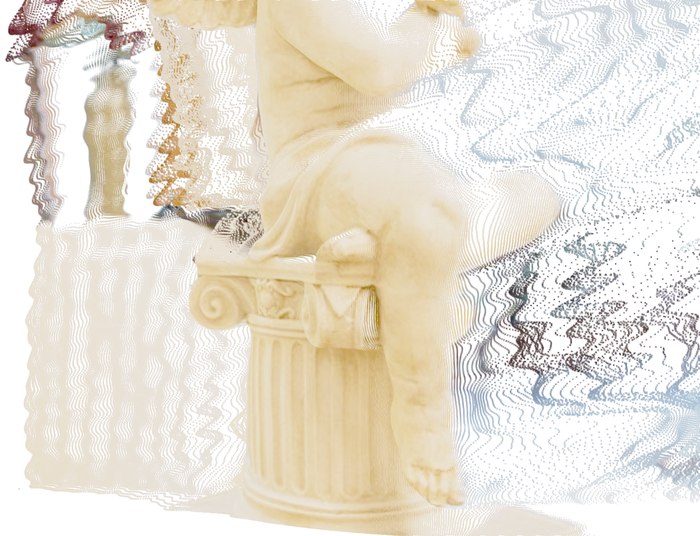} &
    \includegraphics[width=0.244\textwidth]{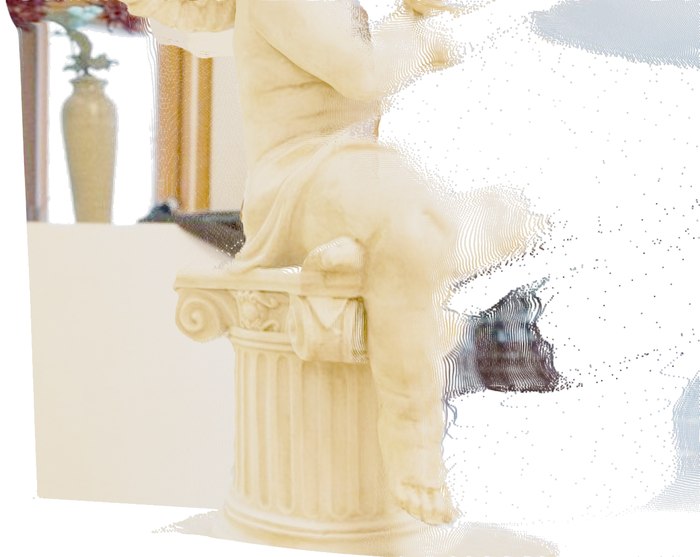} &
    \includegraphics[width=0.244\textwidth]{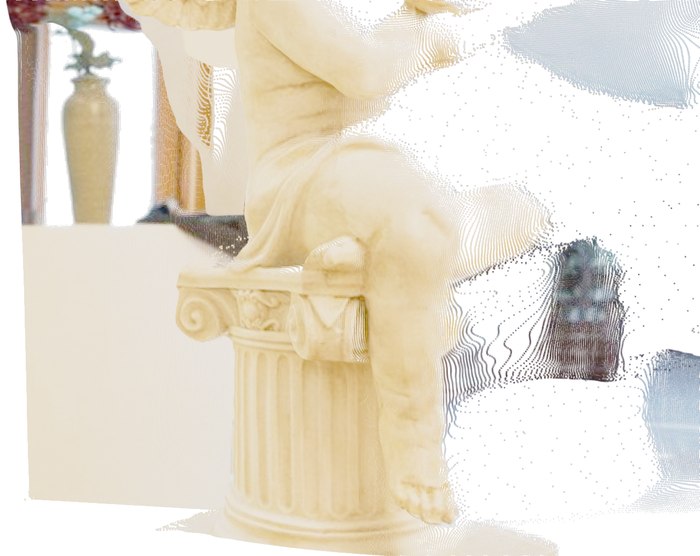} \\[2pt]
  \end{tabular}
  \caption{Point-cloud comparison on two NYU-v2 scenes and one BlendedMVS scene (top to bottom), rendered from a $14^\circ$ offset view. DUSt3R and one-step JUSt3R exhibit streaks and stretched surfaces near depth boundaries, where the conditional mean falls between surfaces; multi-step inference from the same JUSt3R checkpoint separates them.}
  \label{fig:just3r-cloud-comparison}
\end{figure}

\begin{figure}[t]
  \centering
  \begin{minipage}[c]{0.2\textwidth}
    \includegraphics[width=\linewidth]{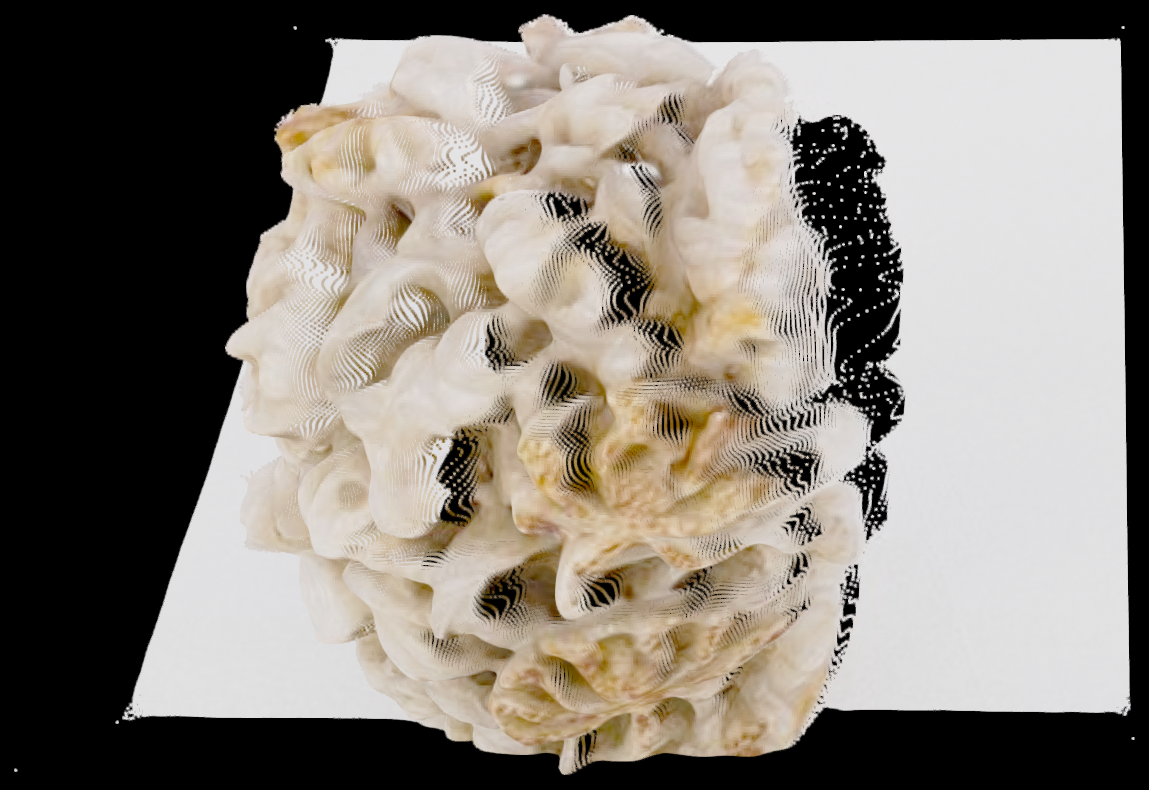}\\[1.5mm]
    \includegraphics[width=\linewidth]{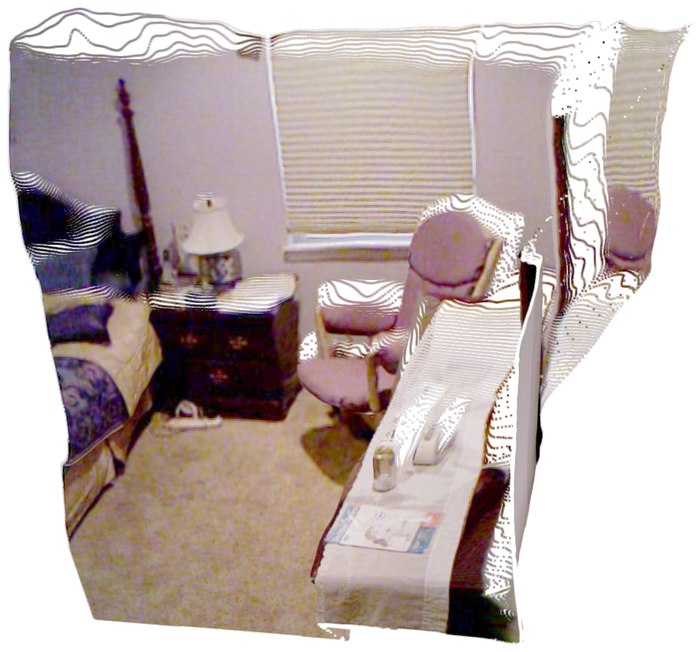}
  \end{minipage}\hfill
  \begin{minipage}[c]{0.7423\textwidth}
    \includegraphics[width=\linewidth]{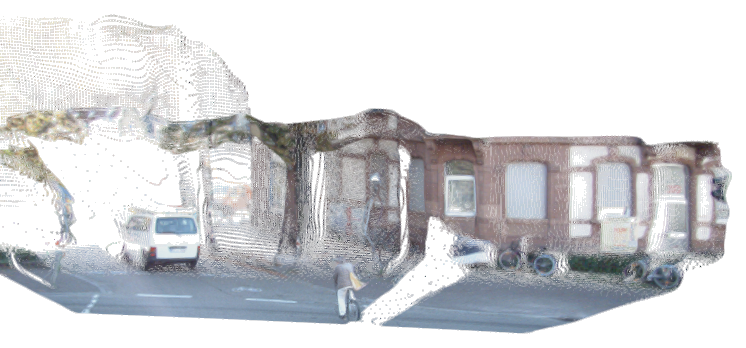}
  \end{minipage}
  \caption{Pointmaps from JUSt3R$_{50}$ on object, indoor, and outdoor inputs.}
  \label{fig:dust3r-clouds}
  \vspace{-6mm}
\end{figure}

\paragraph{Qualitative comparison.}
Figure~\ref{fig:just3r-cloud-comparison} compares DUSt3R with one-, ten-, and fifty-step JUSt3R on two NYU-v2 scenes and one BlendedMVS scene.
DUSt3R and JUSt3R$_1$ produce streaks and stretched surfaces around furniture, wall openings, and the carved column, whereas multi-step predictions separate these surfaces cleanly while preserving the overall layout.
Figure~\ref{fig:dust3r-clouds} shows fifty-step outputs across object, indoor, and outdoor inputs, and Appendix~\ref{app:dust3r-gallery} provides further comparisons.

\paragraph{Quantitative comparison.}
Table~\ref{tab:dust3r} reports five zero-shot depth benchmarks.
\emph{The one-step endpoint is a competitive direct predictor.}
With a single decoder pass, JUSt3R$_1$ improves AbsRel over the retrained DUSt3R on four of the five datasets and improves $\delta_{1.25}$ on four. Therefore training along the flow does not cost direct-prediction accuracy.
\emph{More steps act at boundaries.}
Moving from JUSt3R$_1$ to JUSt3R$_{10}$ with the same checkpoint lowers the veil rate on all four datasets with dense ground truth, lowers the flying-pixel rate on all five (by 16--43\%), and raises BF1 on three of four.
Pointwise accuracy moves in the opposite direction but only slightly.
This asymmetry is what a narrow posterior with localized ambiguity predicts. The interior, which dominates AbsRel, was already close to any sample, while the boundaries, where the mean falls between surfaces, are where committing to one surface matters.
It mirrors the image-space perception--distortion trade-off~\citep{blau2018perception}, and contrasts with sparse-context NVS, where a broad posterior makes the same trade-off global.

\subsection{Empirical Takeaways}
\label{sec:takeaways}

Read together, the three settings support the formulation in Section~\ref{sec:revisit} from complementary angles. The toy experiment checks them against exact references, and the two 3D tasks show how they surface in real output spaces with very different posteriors.

\paragraph{Reconstruction is the one-step endpoint of the flow.}
In the toy, one Euler step reproduces the analytic conditional mean as closely as a dedicated feed-forward regressor, despite the added noise.
JUSt3R$_1$ does the same on a real backbone, improving AbsRel over the retrained DUSt3R on four of five benchmarks.
Conversion to flow thus does not trade away reconstruction.

\paragraph{Committing to a solution trades distortion for realism.}
With more steps, the models commit to single solutions: toy samples return to the data manifold, JiT-LVSM halves FID on sparse-context DL3DV, and JUSt3R removes veils and flying pixels.
In each case, pointwise distortion rises (up to $1.8$\,dB PSNR in NVS, $0.013$ AbsRel in geometry).
The mean fails alike across output spaces, averaging incompatible explanations into points inside the circle, blur in images, and points floating between surfaces.
Hence PSNR or AbsRel favor the conditional mean and understate what generation contributes, while perceptual and structural measures hide what it costs; a fair comparison needs both.

\paragraph{Posterior width sets the price of commitment.}
The trade-off grows with posterior width, which has two sources.
The first is context: as reliability rises in the toy, samples become both realistic and correct, and the PSNR cost of JiT-LVSM under sparse context turns into a gain under rich context.
The second is the task: pointmaps are ambiguous mainly at depth discontinuities, so multi-step inference acts locally, whereas sparse-context NVS is ambiguous over large regions.
Width also sets the risk of generation: under a broad posterior, a committed sample can be realistic but wrong, as in the toy's fixed-mode panels and sparse-context NVS.
Generation does not substitute for missing evidence; it chooses among the completions the evidence leaves open.

\paragraph{Denoising depth is a test-time choice.}
Because one checkpoint serves both roles, the number of steps can be chosen at inference without retraining.
A single step gives the lower-distortion estimate that downstream measurement or alignment may prefer; more steps give sharper, more committed outputs at additional decoder cost, which suits applications that value realism or clean structure.
Which choice is right depends on how strongly the observations constrain the scene, and since that constraint varies across inputs and even across regions of one output, the choice need not be global.

\section{Conclusion}
\label{sec:conclusion}

From feed-forward to flow, the gap between reconstruction and generation is smaller than it appears.
We present a unified flow-based formulation in which direct reconstruction occupies the single-step noise endpoint and conditional generation composes state-dependent clean-target predictions along the flow.
This relationship explains why feed-forward methods can fail despite strong distortion scores: under squared error, their conditional-mean optimum may average incompatible explanations into an output that resembles none of them.
A controlled toy experiment exposes this bias, while JiT-LVSM and JUSt3R connect the same perspective to sharper novel views and cleaner geometric structure.
Together, these results make unification concrete across appearance and geometry, using familiar reconstruction backbones and different denoising configurations.
They also reveal a useful consequence: multi-pass flow inference supplies additional test-time computation, allowing a fixed model to express detail that one-step prediction can average away.
Context constrains that detail; distortion alone does not tell us whether it is useful.

\paragraph{Limitations and future directions.}
Our experiments cover NVS and geometry prediction at a fixed set of denoising depths.
Larger feed-forward architectures such as VGGT~\citep{wang2025vggt}, and the choice of depth per input or spatial region, offer natural extensions of the formulation.
The analysis concerns direct task-loss prediction and flow matching; its one-step endpoint result does not preclude one-step generators trained with different objectives.
Recent consistency models~\citep{song2023consistency}, mean-flow models~\citep{geng2025meanflow,lu2026pixelmeanflow}, and drifting models~\citep{deng2026drifting} motivate a further direction: retaining the benefits of generative inference while reducing the computation needed to obtain them.

\section*{AI Use Statement}
Generative AI tools were used to improve the readability of the manuscript and to assist with LaTeX formatting. All AI-assisted content was reviewed and revised by the authors.
\section*{Reproducibility Statement}
Dataset construction, preprocessing, hyperparameters, training, evaluation and additional implementation details are documented in the main paper and appendix. Code, dataset and pretrained checkpoints will be released upon publication.
\section*{Ethics Statement}
The work is a methodological and empirical study of reconstruction and generation behavior. We are not aware of any
direct ethical concerns arising from this work.
{
    \small
    \bibliographystyle{references}
    \bibliography{references}
}
\newpage
\appendix

\section{Related Work}
\label{sec:related}

\paragraph{Feed-forward 3D prediction.}
Feed-forward 3D systems replace per-scene optimization with direct prediction from images.
Early generalizable radiance-field and image-based rendering models predict neural scene representations or target-view radiance from sparse inputs~\citep{yu2021pixelnerf,wang2021ibrnet,chen2021mvsnerf}.
Recent Gaussian and transformer models further scale this direct-prediction paradigm to sparse-view rendering, view synthesis, cameras, depth, pointmaps, and tracks~\citep{charatan2024pixelsplat,chen2024mvsplat,jin2025lvsm,jiang2025rayzer,wang2025less, wang2024dust3r,leroy2024grounding,wang2025vggt}.
These methods establish feed-forward prediction as a strong default for reconstruction. We place this default within a unified flow formulation to explain its successes, its failures under ambiguity, and its relationship to conditional generation.

\paragraph{Denoising targets.}
Diffusion and score-based models generate samples through iterative denoising~\citep{ho2020denoising,song2020denoising,song2020score,karras2022elucidating}, and flow matching gives a related continuous transport formulation~\citep{lipman2022flow}.
Modern architectures such as diffusion transformers make the denoiser itself increasingly general-purpose~\citep{peebles2023scalable,yu2025pixeldit}.
The denoising target is also a modeling choice: denoisers may predict noise, velocity, scores, or clean data.
Recent pixel-space work revives direct clean-target $x_0$ prediction~\citep{li2025back,yu2025pixeldit}.
Our formulation uses the noise-to-data flow-matching path throughout: clean-target prediction induces its velocity field, connecting the feed-forward endpoint with multi-step generative inference.

\paragraph{Generative and hybrid 3D.}
Generative models have been used to predict ambiguous 3D signals, including novel views, 3D structure, and depth~\citep{watson2022novel,liu2023zero,chan2023generative,ye2024consistent,kani2023upfusion,ke2024repurposing}.
Diffusion priors have also been lifted into 3D reconstruction and generation through distillation, optimization, or reconstruction regularization~\citep{poole2022dreamfusion,zhou2023sparsefusion,wu2024reconfusion}.
More recent systems combine reconstruction and generation priors with tokenized, autoregressive, or diffusion components~\citep{xiang2026rng,chang2025reconviagen,le2025umami,asim2026scenetok}.
These works, together with recent spatial world models such as Atlas~\citep{worldlabs2026atlas}, show the practical relevance of bringing reconstruction and generation together.
We study their shared formulation: reconstruction and generation emerge from different denoising configurations of one flow-based prediction interface, instantiated in both NVS and geometry.

\paragraph{Uncertainty and posterior means.}
The theory of squared-error prediction and Bayesian decision making identifies posterior or conditional means as optimal point estimates~\citep{berger1985statistical,hastie2009elements}.
Posterior-mean denoising also has a classical score-based characterization through Tweedie's formula~\citep{efron2011tweedie}.
Diffusion inverse-problem methods further separate posterior-mean restoration from posterior sampling~\citep{kawar2022denoising,chung2022diffusion}.
The perception--distortion literature formalizes why mean-like estimates and perceptually realistic samples can occupy different points on a trade-off curve~\citep{blau2018perception,blau2019rethinking,freirich2021theory}.
We connect these established statistical ideas to a unified empirical account of 3D prediction. The toy experiment makes conditional averaging visible, and the NVS and geometry experiments show why pointwise distortion alone can miss the benefits of flow inference.

\section{Flow Formulation: Statements and Proofs}
\label{app:proof}

We use the path and clean-target interface of Section~\ref{sec:revisit}, with $\epsilon$ independent of $(X_0,C)$. Conditional distributions and risks are assumed well defined, with measurable optimal actions in an unrestricted prediction class. Squared-error claims assume finite second moments. Endpoint results allow general task losses; exact flow transport uses squared error.

\subsection{Endpoint Prediction and Conditional Risk}
\label{app:endpoint-equivalences}

For positive weights $w(t)$ and $t<1$, flow matching is a weighted clean-target regression:
\begin{equation}
  \mathcal L_{\mathrm{FM}}(\theta)
  =\mathbb E\!\left[w(t)\|v_\theta(X_t,t,C)-(X_0-\epsilon)\|_2^2\right]
  =\mathbb E\!\left[\frac{w(t)}{(1-t)^2}\|f_\theta(X_t,t,C)-X_0\|_2^2\right].
  \label{eq:flow-matching-risk}
\end{equation}
More generally, let $a$ include the clean prediction and any auxiliary outputs, and let $\ell_t$ be their complete loss. The aggregate risk and its pointwise optimal actions are
\begin{align}
  \mathcal R_\nu(a)
  &:=\mathbb E_{t\sim\nu,(X_0,C),\epsilon}
  [\ell_t(X_0,a(X_t,t,C))],
  \label{eq:general-x0-risk}\\
  a_t^\star(x,c)&\in\mathcal B_t(x,c)
  :=\arg\min_a\mathbb E[\ell_t(X_0,a)\mid X_t=x,C=c],
  \label{eq:path-bayes-action}
\end{align}
where $t$ is independent of the other variables. Write $q_t$ for the clean-target readout of $a_t^\star$.

\begin{proposition}[Prediction from an uninformative state]
\label{prop:collapse}
Define
\begin{equation}
  \mathcal B_\ell(c):=\arg\min_a\mathbb E[\ell(X_0,a)\mid C=c].
  \label{eq:context-bayes-set}
\end{equation}
If $S\perp X_0\mid C$, then almost surely
\begin{equation}
  \arg\min_a\mathbb E[\ell(X_0,a)\mid S=s,C=c]=\mathcal B_\ell(c).
  \label{eq:uninformative-source-set}
\end{equation}
A unique optimum ignores $S$; otherwise $S$ can only select among equally optimal actions.
\end{proposition}
\paragraph{Proof.}
Conditional independence gives, for every admissible $a$,
\begin{equation}
  \mathbb E[\ell(X_0,a)\mid S=s,C=c]
  =\mathbb E[\ell(X_0,a)\mid C=c].
  \label{eq:proof-identical-risks}
\end{equation}
The conditional risks, and hence their minimizers, coincide. $\square$

\begin{proposition}[When an optimal one-step predictor cannot sample]
\label{prop:cannot-generate}
Let $\pi_X$ extract the clean prediction and set $\mathcal Q_\ell(c)=\{\pi_X(a):a\in\mathcal B_\ell(c)\}$. Under Proposition~\ref{prop:collapse}, every optimal predictor's conditional output law is supported on $\overline{\mathcal Q_\ell(c)}$. It cannot match a target law assigning positive mass outside this set. In particular, a unique clean readout cannot sample a non-degenerate target law.
\end{proposition}
\paragraph{Proof.}
An optimal action lies in $\mathcal B_\ell(c)$ almost surely, so its clean readout lies in $\mathcal Q_\ell(c)$. $\square$

\paragraph{Reconstruction endpoints.}
LVSM's fixed target-side input and Noised-LVSM's independent noise both satisfy Proposition~\ref{prop:collapse}. Their unrestricted optima therefore agree under the same image loss. Likewise, for a matched complete geometry loss,
\begin{equation}
  \underbrace{\arg\min_a\mathbb E[\ell_{\mathrm{geo}}(X_0,a)\mid\epsilon=e,C=c]}_{\text{JUSt3R endpoint}}
  =\underbrace{\arg\min_a\mathbb E[\ell_{\mathrm{geo}}(X_0,a)\mid C=c]}_{\text{DUSt3R endpoint}}.
  \label{eq:dust3r-just3r-equivalence}
\end{equation}
The action may include confidence and coupled scale normalization. This identity requires the same loss on both sides; it does not equate a norm-loss optimum with a squared-error flow optimum.

For a continuous timestep distribution, $t=0$ has zero mass in Eq.~\eqref{eq:general-x0-risk}. The endpoint statements concern its conditional-risk minimizer. They describe the trained endpoint when it receives positive training mass, or when the predictor and its conditional optimum extend continuously to $t=0$.

\subsection{One-Step Prediction and Denoising Depth}

The clean readout induces a velocity and Euler update
\begin{align}
  \widetilde v_t(x,c)&=\frac{q_t(x,c)-x}{1-t},
  \label{eq:task-loss-velocity}\\
  \widetilde X_{t_{j+1}}
  &=\frac{1-t_{j+1}}{1-t_j}\widetilde X_{t_j}
  +\frac{t_{j+1}-t_j}{1-t_j}q_{t_j}(\widetilde X_{t_j},C),
  \label{eq:x0-euler}
\end{align}
for $0=t_0<\cdots<t_K=1$ and $\widetilde X_{t_0}=\epsilon$.

\begin{proposition}[One-step Euler endpoint]
\label{prop:one-step-endpoint}
One Euler step returns $\widetilde X_1=q_0(\epsilon,C)$. For an optimal endpoint action, this lies in $\mathcal Q_{\ell_0}(C)$. A unique clean readout is independent of $\epsilon$ given $C$; under squared error, it is $\mathbb E[X_0\mid C]$.
\end{proposition}
\paragraph{Proof.}
Set $t_0=0,t_1=1$ in Eq.~\eqref{eq:x0-euler}, then apply Proposition~\ref{prop:collapse}. $\square$

For $0<t_k<1$, the same recurrence unrolls to
\begin{equation}
  \begin{aligned}
    \widetilde X_{t_k}&=(1-t_k)\epsilon+t_k\bar q_k,
    \qquad \bar q_k=\sum_{j<k}\lambda_j^{(k)}q_{t_j}(\widetilde X_{t_j},C),\\
    \lambda_j^{(k)}&=\frac{1-t_k}{t_k}\frac{t_{j+1}-t_j}{(1-t_j)(1-t_{j+1})}.
  \end{aligned}
  \label{eq:unrolled-state}
\end{equation}
The nonnegative weights sum to one by telescoping. Unlike a training state, this state contains predictions in place of the true target.

\begin{proposition}[State dependence beyond one step]
\label{prop:denoising-depth}
If $q_0(C)$ is unique, a two-step schedule $0<s<1$ gives
$\widetilde X_s=(1-s)\epsilon+s q_0(C)$ and $\widetilde X_1=q_s(\widetilde X_s,C)$.
For non-degenerate noise, the intermediate state is non-degenerate given $C$; the output varies exactly when $q_s(\cdot,C)$ is not almost surely constant under that state's law.
\end{proposition}
\paragraph{Proof.}
Apply Eq.~\eqref{eq:x0-euler} twice. The first update retains the nonzero noise coefficient $1-s$; the second evaluates $q_s$ on this state. $\square$

Additional steps thus permit state-dependent predictions. Exact sampling further requires the appropriate velocity and integration, as follows.

\subsection{Ideal Conditional Flow Transport}
\label{app:ideal-flow}

Fix $C=c$, write $\rho_t=\mathcal L(X_t\mid C=c)$, and set $U_t=X_0-\epsilon$. Define
\begin{equation}
  v_t^\star(x,c)=\mathbb E[U_t\mid X_t=x,C=c],\qquad
  m_t(x,c)=\mathbb E[X_0\mid X_t=x,C=c].
  \label{eq:ideal-flow-definitions}
\end{equation}
The squared-error optimum is $v_t^\star$. Since $X_0-\epsilon=(X_0-X_t)/(1-t)$,
\begin{equation}
  v_t^\star(x,c)=\frac{m_t(x,c)-x}{1-t},\qquad t<1.
  \label{eq:ideal-x0-field}
\end{equation}

\begin{proposition}[Transport by the conditional mean field]
\label{prop:ideal-transport}
Assume finite first moments. Then $\rho_t$ satisfies the weak continuity equation
\begin{equation}
  \partial_t\rho_t+\nabla\!\cdot(\rho_t v_t^\star)=0,\qquad \rho_0=p_0.
  \label{eq:continuity-equation}
\end{equation}
If, on each $[0,T]$ with $T<1$, the ODE
\begin{equation}
  \frac{\mathrm dZ_t}{\mathrm dt}=v_t^\star(Z_t,c),\qquad Z_0\sim p_0
  \label{eq:ideal-flow-ode}
\end{equation}
admits a unique flow and its continuity equation has a unique probability-measure solution, then $\mathcal L(Z_t\mid c)=\rho_t$ for $t<1$. These laws converge weakly to $p_{\mathrm{data}}(\cdot\mid c)$ as $t\uparrow1$; any almost-sure terminal limit $Z_1$ has that target law.
\end{proposition}
\paragraph{Proof.}
For a smooth compactly supported test function $\varphi$, differentiation and conditional expectation give
\begin{equation}
  \frac{\mathrm d}{\mathrm dt}\mathbb E[\varphi(X_t)\mid c]
  =\mathbb E[\nabla\varphi(X_t)^\top U_t\mid c]
  =\int\nabla\varphi(x)^\top v_t^\star(x,c)\,\rho_t(\mathrm dx).
  \label{eq:weak-continuity-proof}
\end{equation}
Thus the interpolation and ODE marginals solve the same continuity equation with the same initial law. Uniqueness identifies them; $X_t\to X_0$ gives the terminal marginal limit. $\square$

For a different clean readout, the velocity discrepancy is exactly
\begin{equation}
  \widetilde v_t(x,c)-v_t^\star(x,c)
  =\frac{q_t(x,c)-m_t(x,c)}{1-t}.
  \label{eq:field-decomposition}
\end{equation}
Squared-error flow matching gives $q_t=m_t$ at its population optimum. A general task loss need not, and convergence of clean predictions alone does not control the ratio near $t=1$.

\paragraph{Geometry normalization.}
JUSt3R forms its training path using a jointly normalized two-view pointmap, a fixed representative of the target modulo scale. Intermediate sampler updates instead use per-view normalization by default, with the final prediction returned at raw scale. These projections preserve the endpoint risk argument but distinguish the practical sampler from the ideal ODE above.

\subsection{The Value of Additional Context}

\begin{proposition}[Value of information]
\label{prop:context-value}
If $C_{\mathrm{sparse}}=g(C_{\mathrm{rich}})$ and both settings use the same target, action space, and loss, then
\begin{equation}
  \underbrace{\inf_{\delta_{\mathrm{rich}}}
  \mathbb E[\ell(X_0,\delta_{\mathrm{rich}}(C_{\mathrm{rich}}))]}_{R_{\mathrm{rich}}^\star}
  \le
  \underbrace{\inf_{\delta_{\mathrm{sparse}}}
  \mathbb E[\ell(X_0,\delta_{\mathrm{sparse}}(C_{\mathrm{sparse}}))]}_{R_{\mathrm{sparse}}^\star}.
  \label{eq:value-of-information}
\end{equation}
\end{proposition}
\paragraph{Proof.}
Every sparse-context rule is available with rich context by choosing $\delta_{\mathrm{rich}}=\delta_{\mathrm{sparse}}\circ g$. Taking infima proves the inequality. $\square$

This orders expected optimal risks. It does not imply narrower posteriors for every observation or monotonic scores across empirical co-visibility bins.

\section{Toy Experiment Details}
\label{app:toy}

The toy experiment uses a known conditional distribution, so the conditional mean, the posterior over modes, and the correct latent mode are all measurable.
This appendix records the data distribution, the compared predictors, the flow model and its sampler, the evaluation protocol, and full results for Section~\ref{sec:toy}.
The key design choice is that a single flow checkpoint per reliability and seed is evaluated at every step count, so one-step prediction and multi-step sampling are two inference configurations of the same model.

\paragraph{Data distribution.}
We sample a latent mode:
\begin{equation}
  z \sim \mathrm{Uniform}(\{0,\ldots,K-1\}).
  \label{eq:toy-z}
\end{equation}
In the main experiment, $K=8$ mode centers are placed on a circle of radius $2$:
\begin{equation}
  \mu_k = 2\left(\cos(2\pi k/K),\, \sin(2\pi k/K)\right).
  \label{eq:toy-centers}
\end{equation}
The target point is sampled as:
\begin{equation}
  x_0 \mid z=k
  \sim
  \mathcal{N}(\mu_k,\sigma_m^2 I).
  \label{eq:toy-x0}
\end{equation}
We use $\sigma_m=0.15$.
The context is a noisy one-hot observation of the latent mode:
\begin{equation}
  c_r \mid z=k
  \sim
  \mathcal{N}(e_k,\sigma_{\mathrm{ctx}}(r)^2 I),
  \qquad
  r\in[0,1].
  \label{eq:toy-context-supp}
\end{equation}
Here $e_k$ is the one-hot vector for mode $k$.
\begin{equation}
  \sigma_{\mathrm{ctx}}(r)
  =
  \sigma_{\mathrm{sparse}}
  +
  r\left(\sigma_{\mathrm{rich}}-\sigma_{\mathrm{sparse}}\right),
  \qquad
  \sigma_{\mathrm{sparse}}=2.5,\quad
  \sigma_{\mathrm{rich}}=0.08.
  \label{eq:toy-sigma}
\end{equation}
The reliability parameter $r$ controls the context noise and hence the posterior concentration over modes, while leaving the marginal target distribution unchanged.

\paragraph{Analytic references.}
Because the context likelihood is Gaussian around one-hot vertices, the posterior over modes can be computed analytically:
\begin{equation}
  p(z=k\mid c_r)
  =
  \frac{
    \exp\!\left(-\|c_r-e_k\|_2^2/(2\sigma_{\mathrm{ctx}}(r)^2)\right)
  }{
    \sum_j \exp\!\left(-\|c_r-e_j\|_2^2/(2\sigma_{\mathrm{ctx}}(r)^2)\right)
  }.
  \label{eq:toy-posterior}
\end{equation}
The analytic conditional mean is:
\begin{equation}
  \mathbb{E}[x_0\mid c_r]
  =
  \sum_{k=0}^{K-1}p(z=k\mid c_r)\mu_k.
  \label{eq:toy-mean-supp}
\end{equation}
This is the reference plotted as \texttt{mean} in Figure~\ref{fig:toy}.
We also use an exact posterior sampler, \texttt{exact}, which draws $z\sim p(z\mid c_r)$ from Eq.~\eqref{eq:toy-posterior} and then $x_0\sim\mathcal{N}(\mu_z,\sigma_m^2 I)$.
It is the ideal behavior of a conditional generative model and serves as the reference for multi-step sampling.

\paragraph{Compared predictors.}
\texttt{gt} is sampled from the true target distribution.
\texttt{feed-forward} is a deterministic predictor $f(c_r)\rightarrow x_0$ trained with $\mathbb{E}\|f(c_r)-x_0\|_2^2$.
The flow model is a clean-target predictor $f_\theta(x_t,t,c_r)$ on the noise-to-data path $x_t=(1-t)\epsilon+tx_0$ of Section~\ref{sec:revisit}, with $\epsilon\sim\mathcal{N}(0,I)$.
It is trained with the clean-target form of the flow-matching loss,
\begin{equation}
  \mathcal{L}_{\mathrm{flow}}
  =
  \mathbb{E}\!\left[w(t)\,\|f_\theta(x_t,t,c_r)-x_0\|_2^2\right],
  \qquad
  w(t)=\frac{1}{(1-\min(t,0.95))^2},
  \label{eq:toy-flow-loss}
\end{equation}
which equals the squared error of the induced velocity $(f_\theta-x_t)/(1-t)$ against $x_0-\epsilon$ .
Time is sampled as $t=0$ with probability $0.1$ and $t\sim\mathcal{U}[0,0.95]$ otherwise, so the one-step endpoint receives positive training mass rather than being reached by extrapolation.

Both learned models use an MLP trunk of hidden width $256$ and depth $4$, AdamW with learning rate $10^{-3}$, weight decay $10^{-4}$, and batch size $512$, and one model of each kind is trained per seed and reliability.
In the flow model, the context (a 2-layer MLP) and time (128 sinusoidal features of $1000t$ followed by a 2-layer MLP) modulate every hidden layer through zero-initialized FiLM, and the raw context is also concatenated to the input.
This conditioning pathway brings the flow model to 1.09M parameters, against 0.13M for \texttt{feed-forward}.
\texttt{feed-forward} is trained for $12{,}000$ steps and the flow model for $36{,}000$; longer training (72k steps) and a wider trunk (512) gave no further improvement.

\paragraph{Sampling.}
We integrate the induced velocity with Euler steps on the uniform grid $t_i=i/k$, $i=0,\ldots,k$, starting from $x_{t_0}=\epsilon$.
The final step lands exactly on the clean prediction, so $k=1$ returns $f_\theta(\epsilon,0,c_r)$, the one-step endpoint.
The network's time input is clamped to its training range ($t\le0.95$), which affects only schedules with $k>20$.
All step counts use the same checkpoint; \texttt{flow-1} and \texttt{flow-15} in Figure~\ref{fig:toy} denote $k=1$ and $k=15$.

\paragraph{Evaluation protocol.}
The main sweep uses $r=0.0,0.1,\ldots,1.0$ and five random seeds.
Metrics are evaluated with 16{,}384 samples per seed for $k=1$ and $k=15$, and with 4{,}096 samples per seed for the step sweep $k\in\{1,2,3,5,10,15,25,50\}$.
For a predicted point $\hat{x}$, we report per-coordinate distortion $\mathbb{E}[\|\hat{x}-x_0\|_2^2/2]$, per-coordinate conditional-mean error $\mathbb{E}[\|\hat{x}-\mathbb{E}[x_0\mid c_r]\|_2^2/2]$, nearest-mode distance $\mathbb{E}[\min_k\|\hat{x}-\mu_k\|_2]$, and mode consistency $\Pr[\arg\min_k\|\hat{x}-\mu_k\|_2=z]$.
Diversity is the mean trace of the empirical covariance of eight samples drawn with independent noise for the same context; \texttt{feed-forward} is deterministic, so its diversity is $0$ by construction.
For \texttt{exact}, distortion and diversity are both close to twice its conditional-mean error, as expected for a posterior sampler.

\paragraph{Quantitative results.}
Table~\ref{tab:toy-endpoints} reports the reliabilities shown in Figure~\ref{fig:toy} together with $r=0$, as mean$\pm$std over five seeds.
Low reliability tests whether a predictor follows the posterior mean or samples from the data manifold; high reliability tests whether sampling becomes mode-consistent once the context identifies the latent mode.

\begin{table}[t]
  \centering
  \scriptsize
  \setlength{\tabcolsep}{2.5pt}
  \caption{Toy metrics, mean$\pm$std over five seeds. One flow checkpoint per $(r,\text{seed})$ is evaluated with one Euler step (\texttt{flow-1}) and with 15 (\texttt{flow-15}). \texttt{mean} is the analytic $\mathbb{E}[x_0\mid c]$ and \texttt{exact} the exact posterior sampler. One-step predictors match \texttt{mean}; the 15-step sampler matches the mode selection of \texttt{exact}. CM error: per-coordinate squared error to $\mathbb{E}[x_0\mid c]$; mode dist.: distance to the nearest mode center; mode cons.: fraction of predictions nearest the true mode.}
  \vspace{1mm}
  \label{tab:toy-endpoints}
  \resizebox{\linewidth}{!}{
  \begin{tabular}{llccccc}
    \toprule
    Reliability & Method & CM error $\downarrow$ & Mode dist. $\downarrow$ & Mode cons. $\uparrow$ & Distortion $\downarrow$ & Diversity \\
    \midrule
    $r=0.0$ & \texttt{exact} & $1.98173{\pm}0.00244$ & $0.18787{\pm}0.00069$ & $0.14550{\pm}0.00360$ & $3.95202{\pm}0.02895$ & $3.95706{\pm}0.00373$ \\
    $r=0.0$ & \texttt{mean} & $0.00000{\pm}0.00000$ & $1.75274{\pm}0.00107$ & $0.16716{\pm}0.00050$ & $1.97979{\pm}0.00252$ & -- \\
    $r=0.0$ & \texttt{feed-forward} & $0.00136{\pm}0.00036$ & $1.74957{\pm}0.01735$ & $0.16368{\pm}0.00349$ & $1.98199{\pm}0.00303$ & -- \\
    $r=0.0$ & \texttt{flow-1} & $0.00256{\pm}0.00084$ & $1.76830{\pm}0.01112$ & $0.16583{\pm}0.00163$ & $1.98313{\pm}0.00390$ & $0.00091{\pm}0.00031$ \\
    $r=0.0$ & \texttt{flow-15} & $1.86915{\pm}0.05846$ & $0.19175{\pm}0.01137$ & $0.14235{\pm}0.00379$ & $3.85215{\pm}0.05699$ & $3.72978{\pm}0.11580$ \\
    \midrule
    $r=0.4$ & \texttt{exact} & $1.90867{\pm}0.00603$ & $0.18853{\pm}0.00102$ & $0.17390{\pm}0.00179$ & $3.83301{\pm}0.02287$ & $3.80880{\pm}0.00182$ \\
    $r=0.4$ & \texttt{mean} & $0.00000{\pm}0.00000$ & $1.58902{\pm}0.00148$ & $0.20696{\pm}0.00241$ & $1.90754{\pm}0.00181$ & -- \\
    $r=0.4$ & \texttt{feed-forward} & $0.00218{\pm}0.00046$ & $1.59005{\pm}0.01717$ & $0.20934{\pm}0.00445$ & $1.90542{\pm}0.00610$ & -- \\
    $r=0.4$ & \texttt{flow-1} & $0.00424{\pm}0.00175$ & $1.60486{\pm}0.01325$ & $0.20835{\pm}0.00383$ & $1.90896{\pm}0.00876$ & $0.00165{\pm}0.00029$ \\
    $r=0.4$ & \texttt{flow-15} & $1.79413{\pm}0.03077$ & $0.18713{\pm}0.00657$ & $0.17341{\pm}0.00172$ & $3.69475{\pm}0.03906$ & $3.57342{\pm}0.06062$ \\
    \midrule
    $r=0.9$ & \texttt{exact} & $0.25936{\pm}0.00326$ & $0.18791{\pm}0.00071$ & $0.89568{\pm}0.00192$ & $0.51804{\pm}0.00750$ & $0.52255{\pm}0.00980$ \\
    $r=0.9$ & \texttt{mean} & $0.00000{\pm}0.00000$ & $0.17002{\pm}0.00114$ & $0.90859{\pm}0.00194$ & $0.25670{\pm}0.00528$ & -- \\
    $r=0.9$ & \texttt{feed-forward} & $0.00441{\pm}0.00079$ & $0.19809{\pm}0.00603$ & $0.90739{\pm}0.00267$ & $0.26203{\pm}0.00810$ & -- \\
    $r=0.9$ & \texttt{flow-1} & $0.00788{\pm}0.00272$ & $0.20096{\pm}0.01277$ & $0.90518{\pm}0.00172$ & $0.27021{\pm}0.00454$ & $0.00399{\pm}0.00053$ \\
    $r=0.9$ & \texttt{flow-15} & $0.23120{\pm}0.00654$ & $0.16251{\pm}0.00685$ & $0.89617{\pm}0.00390$ & $0.48006{\pm}0.01744$ & $0.45125{\pm}0.01516$ \\
    \midrule
    $r=1.0$ & \texttt{exact} & $0.02261{\pm}0.00021$ & $0.18841{\pm}0.00096$ & $1.00000{\pm}0.00000$ & $0.04489{\pm}0.00046$ & $0.04504{\pm}0.00015$ \\
    $r=1.0$ & \texttt{mean} & $0.00000{\pm}0.00000$ & $0.00000{\pm}0.00000$ & $1.00000{\pm}0.00000$ & $0.02250{\pm}0.00028$ & -- \\
    $r=1.0$ & \texttt{feed-forward} & $0.00030{\pm}0.00014$ & $0.02132{\pm}0.00592$ & $1.00000{\pm}0.00000$ & $0.02277{\pm}0.00031$ & -- \\
    $r=1.0$ & \texttt{flow-1} & $0.00049{\pm}0.00021$ & $0.02725{\pm}0.00697$ & $1.00000{\pm}0.00000$ & $0.02285{\pm}0.00024$ & $0.00017{\pm}0.00006$ \\
    $r=1.0$ & \texttt{flow-15} & $0.01566{\pm}0.00102$ & $0.15657{\pm}0.00538$ & $0.99999{\pm}0.00003$ & $0.03822{\pm}0.00085$ & $0.02830{\pm}0.00198$ \\
    \bottomrule
  \end{tabular}
  }
\end{table}

\textbf{One-step endpoint.}
Across all eleven reliabilities, the conditional-mean error of \texttt{flow-1} is at most $0.009$ (mean over seeds), the same order as \texttt{feed-forward} (at most $0.0044$) and two orders of magnitude below any posterior sampler.
At $r=0.0$ and $r=0.4$, both one-step predictors lie far from the mode centers (nearest-mode distance above $1.5$, against $0.19$ for data): the conditional mean sits between the modes.

\textbf{Multi-step sampling.}
From the same checkpoint, \texttt{flow-15} returns to the data manifold and matches the mode consistency of \texttt{exact} at every reliability: it is at chance when the context is ambiguous ($0.142$ against $0.146$ at $r=0$), and identical to \texttt{exact} at $r=0.9$ ($0.896$) and $r=1.0$ ($1.000$).
Table~\ref{tab:toy-highr} repeats this comparison on a finer grid near rich context, where the posterior contracts fastest; \texttt{flow-15} follows the exact curve throughout the transition.

\begin{table}[t]
  \centering
  \caption{Mode consistency near rich context (seed 0). \texttt{flow-15} tracks the exact posterior sampler as the posterior contracts.}
  \vspace{1mm}
  \label{tab:toy-highr}
  \begin{tabular}{lccccc}
    \toprule
    & $r=0.80$ & $r=0.85$ & $r=0.90$ & $r=0.95$ & $r=1.00$ \\
    \midrule
    \texttt{exact}   & $0.510$ & $0.689$ & $0.898$ & $0.998$ & $1.000$ \\
    \texttt{flow-15} & $0.519$ & $0.690$ & $0.895$ & $0.997$ & $1.000$ \\
    \bottomrule
  \end{tabular}
\end{table}

\textbf{Within-mode dispersion.}
The one systematic gap is within-mode spread at high reliability: at $r=1.0$, \texttt{flow-15} places every sample on the correct mode but with nearest-mode distance $0.157$ and diversity $0.028$, against $0.188$ and $0.045$ for \texttt{exact}.
To separate learning error from sampling error, we run the same Euler sampler with the Bayes-optimal clean-target predictor $\mathbb{E}[x_0\mid x_t,c_r]$, which is available in closed form for this Gaussian mixture.
It shows the same shortfall (nearest-mode distance $0.151$ and diversity $0.029$ at $k=15$), because the within-mode spread ($\sigma_m=0.15$) is determined at $t\gtrsim0.85$, where a uniform 15-step grid takes only one or two steps.
The gap therefore reflects discretization near the data endpoint, not a failure of the learned predictor or its conditioning. In fact, it narrows with more steps (nearest-mode distance $0.176$ at $k=50$).

\paragraph{Robustness to $K=16$.}
With $16$ modes ($r\in\{0.0,0.5,1.0\}$, three seeds, 8{,}192 samples per seed), the same behavior holds.
The one-step endpoint matches the analytic mean (conditional-mean error $0.002$, $0.005$, and $0.0007$), and at $r=1.0$ the 15-step samples reach mode consistency $0.996$, against $0.991$ for \texttt{exact}; at $r=0.0$ both are at chance.

\section{LVSM and JiT-LVSM Experiment Details}
\label{app:additional}

This section provides implementation details for the LVSM-family experiments used in the main paper.
We organize the details around the experimental protocol, controlled model variants, shared backbone, JiT-LVSM objective, training recipe, inference procedure, and evaluation metrics.
Unless otherwise stated, all LVSM-family variants share the same conditioning inputs and backbone shape, so that the comparisons mainly isolate the effect of target-side source distribution, training objective, and denoising depth.

\paragraph{Datasets, protocols, and context-richness proxies.}
The real-NVS experiments are conducted on RealEstate10K~\citep{ZhouTFFS18} and DL3DV~\citep{ling2024dl3dv}.
We evaluate both interpolation and extrapolation protocols.
In the interpolation protocol, target views remain close to the observed view span.
In the extrapolation protocol, target views are less directly supported by the source images and therefore expose ambiguity more strongly.
The main results report sparse- and rich-context splits and a source-overlap probe based on RoMA~\citep{edstedt2024roma}. These characterize the evidence available to constrain a target view; they are practical context descriptors rather than measurements of conditional entropy.

\paragraph{Controlled model variants.}
We compare three LVSM-family variants.
The LVSM baseline is the original one-step feed-forward reconstruction model.
It uses source RGB and Plucker-ray tokens together with target Plucker-ray tokens, and directly predicts target RGB patches.
Noised-LVSM changes only the target-side source distribution: it replaces blank target-side RGB channels with Gaussian noise while keeping the same one-step $x_0$ prediction objective.
JiT-LVSM instantiates the shared clean-target flow interface with the LVSM backbone and conditioning inputs. It trains on intermediate noise-to-data states with a flow-matching objective and image-space supervision.
During training, the target-side image input is a time-dependent mixture of Gaussian noise and the ground-truth target image.
During inference, JiT-LVSM starts from noise and synthesizes the target view through multi-step denoising.

\paragraph{Shared backbone architecture.}
Unless otherwise stated, LVSM, Noised-LVSM, and JiT-LVSM use the same decoder-only transformer backbone.
Images are resized to $224\times224$ and tokenized with $14\times14$ patches, giving a $16\times16$ patch grid per view.
Each image token has $9$ input channels: RGB plus two $3$-channel ray or camera-direction representations.
The transformer has hidden width $768$, head dimension $64$, and $12$ layers.
We use QK normalization, rotary positional embeddings with frequency $100$, VGGT-style attention, special initialization, and depth initialization.

\paragraph{JiT-LVSM objective and time parameterization.}
JiT-LVSM follows the clean-data prediction interface of JiT while preserving the LVSM conditioning interface.
For each target image $x_0$, we sample Gaussian noise $\epsilon$ and a flow-matching time $t$ uniformly along the noise-to-data path.
The target-side input is the linear interpolation $x_t=t x_0+(1-t)\epsilon$.
The network predicts $\hat{x}_0$ from source-view tokens, target-ray tokens, $x_t$, and a timestep embedding injected into the transformer blocks.
Flow matching is applied through the induced velocity, alongside an image-space task loss on the clean prediction. We describe these components as the $v$-loss and $x$-loss, respectively.
The $x$-loss directly supervises the clean prediction $\hat{x}_0$ with a pixel MSE term and a perceptual term.
The recorded representative configuration uses a pixel MSE weight of $1.0$, a perceptual-loss weight of $0.5$, and an LPIPS-loss weight of $0.0$.
The $v$-loss applies flow matching to the velocity induced by the predicted clean image, following the clean-target parameterization of JiT.
Specifically, the target velocity is $(x_0-x_t)/(1-t)$ and the predicted velocity is $(\hat{x}_0-x_t)/(1-t)$.
This velocity parameterization can become unstable near the data endpoint, so we follow the JiT-style stabilization and use an endpoint cutoff $t_{\epsilon}=0.05$.
The representative configuration weights this $v$-loss by $0.1$.
Training timesteps follow the standard uniform flow-matching schedule with this endpoint cutoff.
Both terms act through the same clean-target prediction. As Eq.~\eqref{eq:flow-matching-risk} shows, the squared velocity loss itself induces a time-dependent weighting of clean-target error; its relative contribution cannot be read from the scalar loss weights alone.

\paragraph{Training setup.}
The representative JiT-LVSM run uses RealEstate10K full-resolution processed training and test lists.
Each training sample contains $2$ input views and $6$ target views, for $8$ total views.
Source views are sampled with frame distance between $25$ and $192$ frames.
We use square center crops with scene scale factor $1.35$.
Training uses mixed precision with fp16 activations, TF32 enabled, and gradient checkpointing every transformer layer.
The per-GPU batch size is $16$ over $8$ GPUs, giving a global batch size of $128$ with no gradient accumulation.
We train for $100$k optimization steps with $3$k warmup steps followed by cosine learning-rate decay.
The peak learning rate is $4\times10^{-4}$.
The optimizer uses AdamW-style moments $(\beta_1,\beta_2)=(0.9,0.95)$, weight decay $0.05$, and gradient clipping at norm $3.0$.
We evaluate and checkpoint every $4$k steps, and visualize $8$ evaluation batches per evaluation.

\paragraph{Inference and compute.}
For multi-step JiT-LVSM evaluation and visual results, we use Heun integration with $10$ denoising steps unless otherwise specified.
The inference time grid follows the standard uniform flow-matching schedule from Gaussian noise to the data endpoint.
The main JiT-LVSM training run used $8$ A100 GPUs for approximately $20$ hours.

\paragraph{Evaluation metrics and qualitative analysis.}
We report PSNR and SSIM for reference-image agreement, together with LPIPS and FID for perceptual and distributional comparison.
These measurements are read alongside qualitative results; none alone fully characterizes agreement with the underlying scene.
Qualitative comparisons are used to expose the main trade-off: feed-forward outputs can be blurrier but more conservative, while multi-step denoising can produce sharper details that may or may not be scene-consistent under sparse context.

\subsection{Target-Side Noise Sanity Check}
\label{app:noised-lvsm}

\begin{wrapfigure}{r}{0.43\linewidth}
  \centering
  \includegraphics[width=\linewidth]{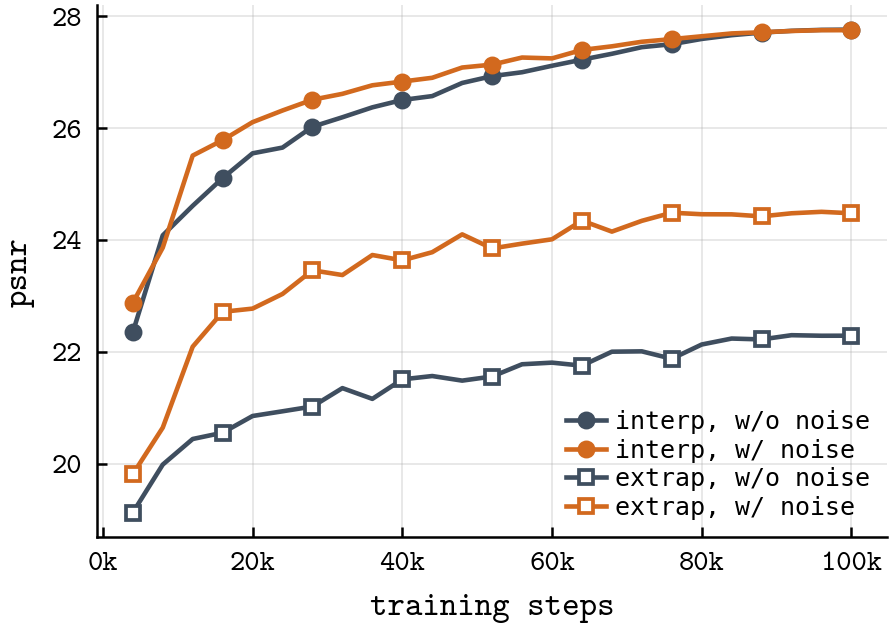}
  \caption{Noised-LVSM on RealEstate10K: extrapolation improves while interpolation remains comparable.}
  \label{fig:source-sanity}
\end{wrapfigure}

Before training along the full flow, we test whether changing only the target-side input affects a one-step predictor. LVSM fills its target-side RGB channels with blanks. Noised-LVSM replaces these blanks with Gaussian noise while retaining the same backbone, context, and one-step clean-image objective. This puts the target-side state at the noise endpoint of the formulation, without adding target-specific information beyond the context.

Figure~\ref{fig:source-sanity} isolates the effect of the target-side input: extrapolation improves with Gaussian noise even though the task remains one-step prediction. This finite-model result complements the population endpoint equivalence. An unchanged optimal-action set does not require identical optimization or generalization in trained networks.
\par\WFclear

\section{DUSt3R and JUSt3R Experiment Details}
\label{app:dust3r}

This section documents the geometry instance of the unified flow formulation: the model, training recipe, evaluation protocol, and boundary metrics used in Section~\ref{sec:dust3r}.

\paragraph{Model.}
JUSt3R uses DUSt3R's encoder, decoder backbone, and output heads~\citep{wang2024dust3r}. Three components are added: (i) a noisy-target embedding that patch-embeds $x_t$, passes it through a two-layer ViT and a zero-initialized projection, and adds the result to
the decoder tokens; (ii) a sinusoidal timestep embedding with a zero-initialized MLP; and (iii) adaLN modulation of each decoder block
(shift/scale/gate on self-attention, cross-attention and MLP), applied externally so that CroCo blocks are unmodified. Gates are initialized to one and shifts/scales to zero, preserving the initial feed-forward computation before the new conditioning branches are learned. Classifier-free guidance dropout is applied to the encoder features with probability $0.1$ during training. The noisy-target module is a two-layer width-$384$ transformer, accompanied by the timestep MLP; these additions are small relative to the ViT-L encoder and twelve-layer decoder. The encoder can be shared across sampling steps. Thus $k$ steps require repeated decoder evaluations over one context encoding, with cost also depending on the solver and guidance settings.

\paragraph{Geometry task loss and the endpoint analysis.}
DUSt3R regresses pointmaps with a confidence-weighted $L_{2,1}$ objective. Both views are expressed in camera view of view~1, write
$X^{v}_{i}\in\mathbb{R}^{3}$ and $\bar X^{v}_{i}\in\mathbb{R}^{3}$ for the predicted and ground-truth coordinates of pixel $i$ in view
$v\in\{1,2\}$, $\mathcal{V}^{v}$ for its valid pixels, and $C^{v}_{i}>0$ for the predicted confidence. The per-pixel regression term is the
Euclidean distance between scale-normalized pointmaps,
\begin{equation}
  \ell(v,i)=\Bigl\lVert
  \tfrac{1}{z}\,X^{v}_{i}-\tfrac{1}{\bar z}\,\bar X^{v}_{i}
  \Bigr\rVert_{2},
  \qquad
  z=\frac{1}{|\mathcal{V}^{1}|+|\mathcal{V}^{2}|}
  \sum_{v}\sum_{i\in\mathcal{V}^{v}}\lVert X^{v}_{i}\rVert_{2},
  \label{eq:l21}
\end{equation}
with $\bar z$ the same average distance computed on the ground truth, so the comparison is scale-invariant. Note that the $\lVert\cdot\rVert_{2}$ is taken to the first power, \emph{not} squared. The full objective then weights each term by its confidence and rewards confident pixels through a $-\alpha\log C$ regularizer, and we get
\begin{equation}
  \mathcal{L}
  =\sum_{v\in\{1,2\}}\frac{1}{|\mathcal{V}^{v}|}
  \sum_{i\in\mathcal{V}^{v}}
  \Bigl(C^{v}_{i}\,\ell(v,i)-\alpha\log C^{v}_{i}\Bigr),
  \qquad \alpha=0.2 .
  \label{eq:confloss}
\end{equation}
The action in this objective contains both the pointmap and confidence, and scale normalization couples its coordinates. Consequently, its Bayes optimum is a joint action; it cannot in general be identified with independent pixelwise geometric medians. The geometric-median characterization applies to a simpler Euclidean norm loss with fixed weights and no prediction-dependent normalization.
This distinction leaves the endpoint result intact: conditional independence makes the full conditional risk the same with or without independent target-side noise (Proposition~\ref{prop:collapse}). The geometry experiments therefore examine the structural behavior of a direct estimator and multi-step flow predictions, without requiring DUSt3R's learned output to equal an analytic mean or median.

\paragraph{Flow-matching formulation.}
JUSt3R predicts a clean, normalized pointmap $\widehat G_0=f_\theta(G_t,t,C)$ from $G_t=(1-t)\epsilon+tG_0$, where $G_0$ is the normalized target defined below. Its velocity is
\begin{equation}
  v_\theta(G_t,t,C)=\frac{\widehat G_0-G_t}{1-t},
  \qquad U_t=G_0-\epsilon.
  \label{eq:geometry-flow-velocity}
\end{equation}
The flow-matching objective supervises this velocity against $U_t$, using the same clean-target interface as JiT-LVSM. Equation~\eqref{eq:confloss} specifies the geometry task criterion used to describe the reconstruction endpoint; it is not a substitute for the complete flow-training objective. The endpoint equality in Eq.~\eqref{eq:dust3r-just3r-equivalence} holds when the complete criterion is matched.

\paragraph{Noise-space normalization.}
Pointmaps have scene-dependent scale (indoor $\sim1$m, outdoor $\sim100$m), so adding $\mathcal{N}(0,I)$ noise in raw coordinates gives inconsistent noise-to-signal ratios. We normalize the target by its average distance from the origin before forming $x_t$, which is the same statistic the loss uses. Therefore, the model operates in a unit-scale space and global alignment recovers metric scale downstream. This fixes the vector-space representative used by the flow interpolation; the pointmap comparison remains scale-invariant. Appendix~\ref{app:proof} discusses the relation between this training normalization and intermediate sampler projections.

\paragraph{Training.}
The training of JUSt3R follows the official DUSt3R three-stage recipe without modification: stage~1 is trained at resolution of $224$ with a linear head ($100$ epochs, $10$ epochs for warmup), stage~2 is trained at resolution $512$ with a linear head ($100$ epochs, $20$ epochs for warmup), stage~3 is trained at resolution of $512$ with a DPT head ($90$ epochs, $15$ epochs for warmup); learning rate $10^{-4}$ throughout. JUSt3R shares the stage-1 initialization and data mixture of DUSt3R.

\paragraph{Evaluation protocol and metrics.}
Following the two-view depth protocol of MonST3R and CUT3R~\citep{zhang2025monst3r,wang2025continuous}, we evaluate on five
monocular-depth benchmarks unseen during training: NYU-v2~\citep{silberman2012indoor}, TUM~\citep{sturm2012benchmark}, Bonn~\citep{palazzolo2019refusion}, Sintel~\citep{butler2012naturalistic}, and KITTI~\citep{geiger2013vision}. Each image is paired with itself, and we average the $z$-coordinate of the  first-view pointmap across the two symmetrized orderings. Before evaluation, each prediction is aligned to the ground truth using a single median scale factor. \textbf{Distortion.} We report the standard pointwise metrics
AbsRel, defined as the mean of $|\hat z-z|/z$, and $\delta_{1.25}$, the fraction of pixels satisfying $\max(\hat z/z,z/\hat z)<1.25$.
\textbf{Geometric realism.} We additionally report three metrics that capture boundary artifacts overlooked by pointwise distortion. \emph{Veil rate} measures predictions that fall into the empty space between two surfaces. We identify ground-truth crossings using $3\times3$ windows whose depth contrast satisfies $(z_{\mathrm{far}}-z_{\mathrm{near}})/z_{\mathrm{near}}>0.1$ and whose valid depths all lie within $0.25(z_{\mathrm{far}}-z_{\mathrm{near}})$ of either extreme, thereby excluding smooth slopes and multi-surface junctions. For each crossing, we compute
\[
t=\frac{\hat z-z_{\mathrm{near}}}
        {z_{\mathrm{far}}-z_{\mathrm{near}}},
\]
and report the fraction for which $0.25<t<0.75$. \emph{Flying-pixel rate} is computed after unprojecting the aligned depth with the dataset intrinsics. At each point $p$, we estimate the surface normal $n(p)$ from the covariance of its $k=16$ nearest 3D neighbours and flag grazing geometry when
\[
\left|\left\langle n(p),\frac{p}{\lVert p\rVert}\right\rangle\right|
<\cos 85^\circ=0.087,
\]
following the ShadowFilter criterion~\citep{pomerleau2013comparing}.
Finally, \emph{BF1} is the tolerance-free, scale-invariant boundary F1 of Depth Pro~\citep{bochkovskiy2025depth}, computed on inverse depth in four directions. Precision and recall require exact agreement between predicted and ground-truth neighbour-pair edges, without dilation or non-maximum suppression, and are averaged over ten ratio thresholds linearly spaced in $[1.05,1.25]$, with weights proportional to the threshold. Ground-truth holes are excluded through neighbour-pair validity masks.

\paragraph{Additional visual comparisons.}
Appendix~\ref{app:dust3r-gallery} compares depth and point clouds on NYU-v2~\citep{silberman2012indoor}, KITTI~\citep{geiger2013vision}, and BlendedMVS~\citep{yao2020blendedmvs} at resolution 512. Each case places its input, reference depth, and DUSt3R/JUSt3R predictions directly above the corresponding point clouds, rendered from a $14^\circ$ offset view. Wide KITTI depth panels use two rows to preserve legibility. These comparisons show how multi-step JUSt3R reduces flying pixels and separates surfaces near occlusion boundaries.

\clearpage
\section{More Visual Results for JiT-LVSM}
\label{app:more-visuals}

\newcommand{\suppimgcell}[1]{\begin{minipage}[t]{0.14\textwidth}\vspace{0pt}\includegraphics[width=\linewidth]{#1}\end{minipage}}
\newcommand{\suppinputcell}[1]{\begin{minipage}[t]{0.0675\textwidth}\vspace{0pt}\includegraphics[width=\linewidth]{fig/picks_dl3dv_random/#1/input_00.jpg}\\[1pt]\includegraphics[width=\linewidth]{fig/picks_dl3dv_random/#1/input_01.jpg}\end{minipage}}
\newcommand{\suppscene}[2]{
  \suppinputcell{#1} &
  \suppimgcell{fig/picks_dl3dv_random/#1/view#2_baseline.jpg} &
  \suppimgcell{fig/picks_dl3dv_random/#1/view#2_jit.jpg} &
  \suppimgcell{fig/picks_dl3dv_random/#1/view#2_gt.jpg}
}
\newcommand{\suppreinputcell}[1]{\begin{minipage}[t]{0.0675\textwidth}\vspace{0pt}\includegraphics[width=\linewidth]{fig/picks_re10k_random/#1/input_00.jpg}\\[1pt]\includegraphics[width=\linewidth]{fig/picks_re10k_random/#1/input_01.jpg}\end{minipage}}
\newcommand{\supprescene}[2]{
  \suppreinputcell{#1} &
  \suppimgcell{fig/picks_re10k_random/#1/view#2_baseline.jpg} &
  \suppimgcell{fig/picks_re10k_random/#1/view#2_jit.jpg} &
  \suppimgcell{fig/picks_re10k_random/#1/view#2_gt.jpg}
}

\begin{figure}[H]
  \centering
  \setlength{\tabcolsep}{0.5pt}
  \renewcommand{\arraystretch}{0.2}
  \makebox[\textwidth][c]{%
  \begin{tabular}{c ccc @{\hspace{5pt}} c ccc}
    {\scriptsize Inputs} &
    {\scriptsize LVSM} & {\scriptsize JiT-LVSM} & {\scriptsize Reference} &
    {\scriptsize Inputs} &
    {\scriptsize LVSM} & {\scriptsize JiT-LVSM} & {\scriptsize Reference} \\
    \suppscene{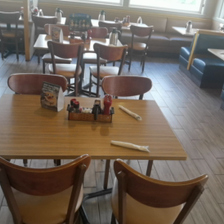}{0190} & \suppscene{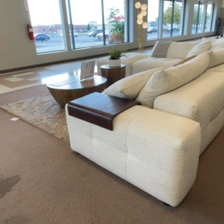}{0150} \\[1pt]
    \suppscene{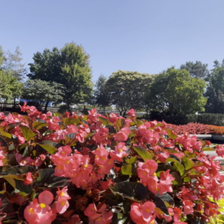}{0167} & \suppscene{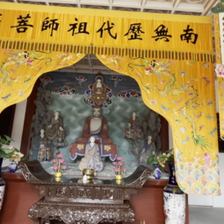}{0098} \\[1pt]
    \suppscene{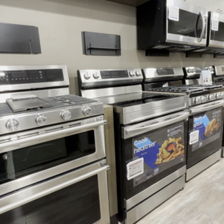}{0201} & \suppscene{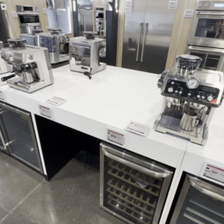}{0014} \\[1pt]
    \suppscene{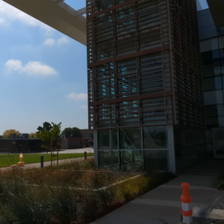}{0150} & \suppscene{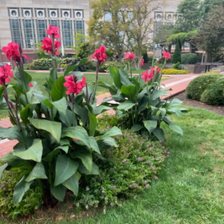}{0306} \\[1pt]
    \suppscene{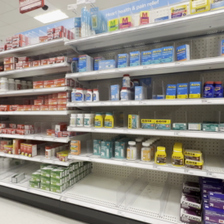}{0263} & \suppscene{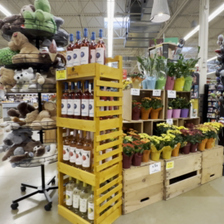}{0211} \\[1pt]
    \suppscene{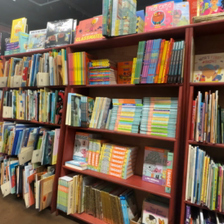}{0221} & \suppscene{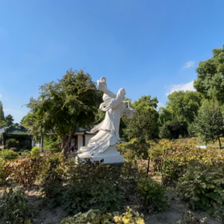}{0019} \\[1pt]
    \suppscene{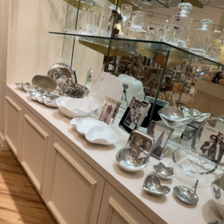}{0122} & \suppscene{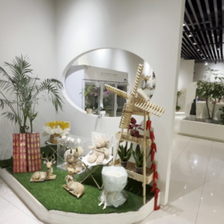}{0150} \\[1pt]
    \suppscene{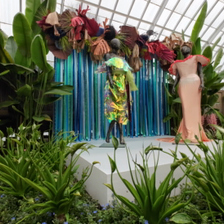}{0187} & \suppscene{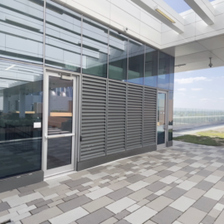}{0164} \\[1pt]
    \suppscene{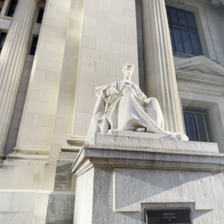}{0230} & \suppscene{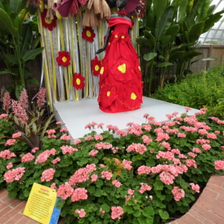}{0101} \\[1pt]
    \suppscene{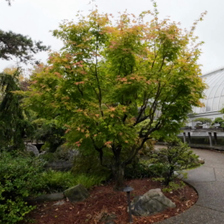}{0190} & \suppscene{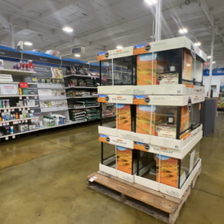}{0189}
  \end{tabular}
  }
  \caption{Random DL3DV visual samples, page 1. Each row shows two randomly selected scenes; each scene includes two input views and one target view, with columns ordered as LVSM, JiT-LVSM, and reference.}
  \label{fig:supp-random-qual-a}
\end{figure}
\clearpage

\begin{figure}[H]
  \centering
  \setlength{\tabcolsep}{0.5pt}
  \renewcommand{\arraystretch}{0.2}
  \makebox[\textwidth][c]{%
  \begin{tabular}{c ccc @{\hspace{5pt}} c ccc}
    {\scriptsize Inputs} &
    {\scriptsize LVSM} & {\scriptsize JiT-LVSM} & {\scriptsize Reference} &
    {\scriptsize Inputs} &
    {\scriptsize LVSM} & {\scriptsize JiT-LVSM} & {\scriptsize Reference} \\
    \suppscene{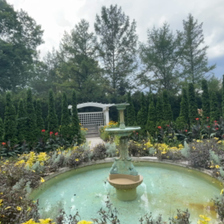}{0100} & \suppscene{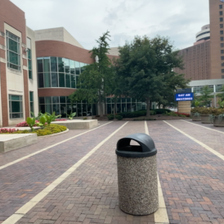}{0192} \\[1pt]
    \suppscene{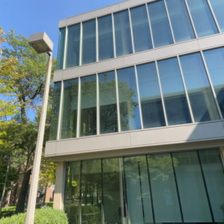}{0076} & \suppscene{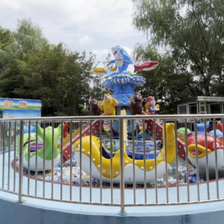}{0228} \\[1pt]
    \suppscene{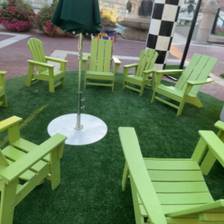}{0227} & \suppscene{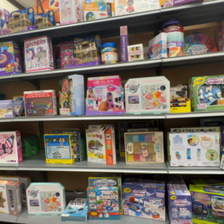}{0210} \\[1pt]
    \suppscene{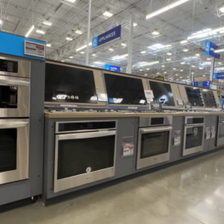}{0268} & \suppscene{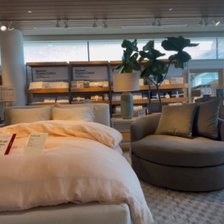}{0212} \\[1pt]
    \suppscene{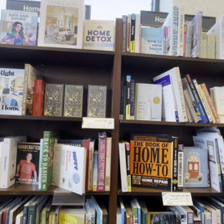}{0215} & \suppscene{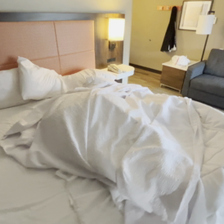}{0236} \\[1pt]
    \suppscene{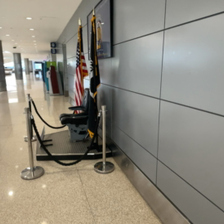}{0109} & \suppscene{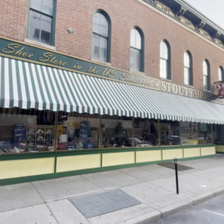}{0046} \\[1pt]
    \suppscene{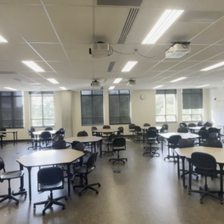}{0099} & \suppscene{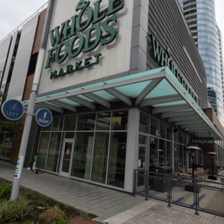}{0112} \\[1pt]
    \suppscene{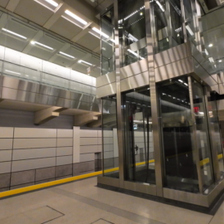}{0115} & \suppscene{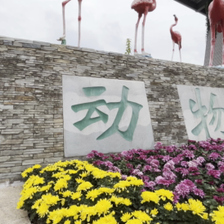}{0280} \\[1pt]
    \suppscene{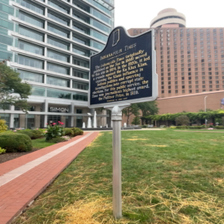}{0201} & \suppscene{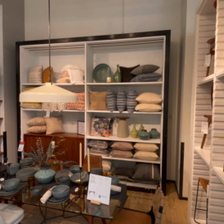}{0117} \\[1pt]
    \suppscene{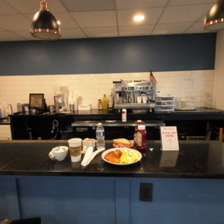}{0073} & \suppscene{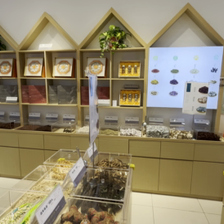}{0101}
  \end{tabular}
  }
  \caption{Random DL3DV visual samples, page 2. Each row shows two randomly selected scenes; each scene includes two input views and one target view, with columns ordered as LVSM, JiT-LVSM, and reference.}
  \label{fig:supp-random-qual-b}
\end{figure}
\clearpage

\begin{figure}[H]
  \centering
  \setlength{\tabcolsep}{0.5pt}
  \renewcommand{\arraystretch}{0.2}
  \makebox[\textwidth][c]{%
  \begin{tabular}{c ccc @{\hspace{5pt}} c ccc}
    {\scriptsize Inputs} &
    {\scriptsize LVSM} & {\scriptsize JiT-LVSM} & {\scriptsize Reference} &
    {\scriptsize Inputs} &
    {\scriptsize LVSM} & {\scriptsize JiT-LVSM} & {\scriptsize Reference} \\
    \suppscene{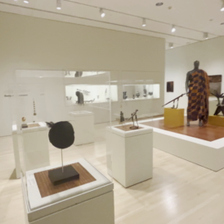}{0105} & \suppscene{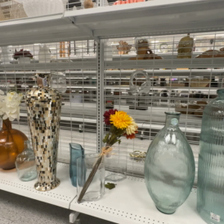}{0206} \\[1pt]
    \suppscene{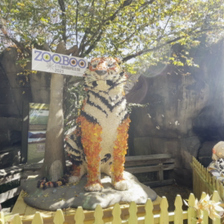}{0049} & \suppscene{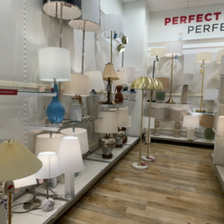}{0055} \\[1pt]
    \suppscene{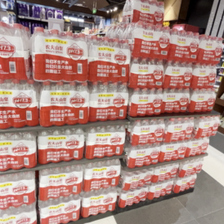}{0186} & \suppscene{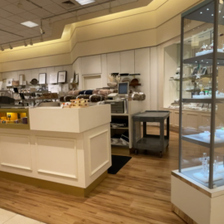}{0076} \\[1pt]
    \suppscene{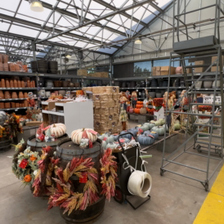}{0178} & \suppscene{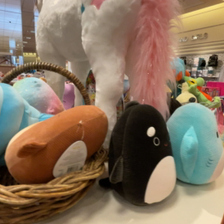}{0321} \\[1pt]
    \suppscene{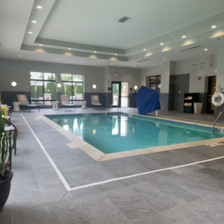}{0073} & \suppscene{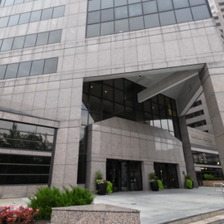}{0185} \\[1pt]
    \suppscene{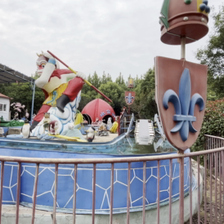}{0202} & \suppscene{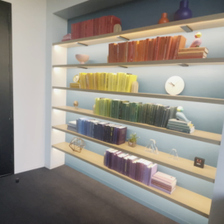}{0087} \\[1pt]
    \suppscene{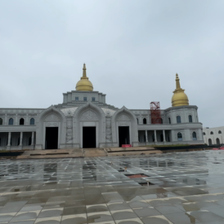}{0124} & \suppscene{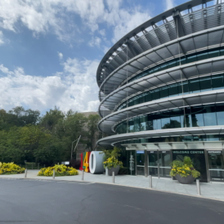}{0257} \\[1pt]
    \suppscene{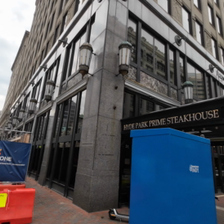}{0059} & \suppscene{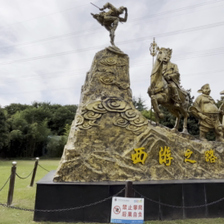}{0257} \\[1pt]
    \suppscene{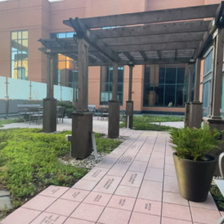}{0017} & \suppscene{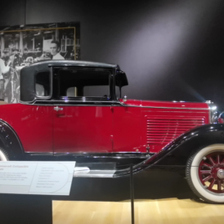}{0273} \\[1pt]
    \suppscene{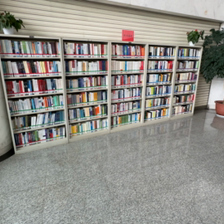}{0182} & \suppscene{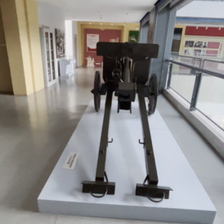}{0178}
  \end{tabular}
  }
  \caption{Random DL3DV visual samples, page 3. Each row shows two randomly selected scenes; each scene includes two input views and one target view, with columns ordered as LVSM, JiT-LVSM, and reference.}
  \label{fig:supp-random-qual-c}
\end{figure}
\clearpage

\begin{figure}[H]
  \centering
  \setlength{\tabcolsep}{0.5pt}
  \renewcommand{\arraystretch}{0.2}
  \makebox[\textwidth][c]{%
  \begin{tabular}{c ccc @{\hspace{5pt}} c ccc}
    {\scriptsize Inputs} &
    {\scriptsize LVSM} & {\scriptsize JiT-LVSM} & {\scriptsize Reference} &
    {\scriptsize Inputs} &
    {\scriptsize LVSM} & {\scriptsize JiT-LVSM} & {\scriptsize Reference} \\
    \supprescene{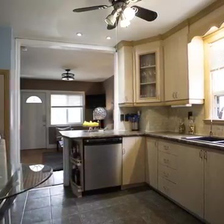}{0051} & \supprescene{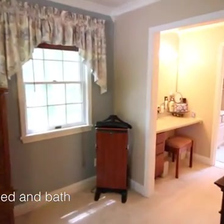}{0036} \\[1pt]
    \supprescene{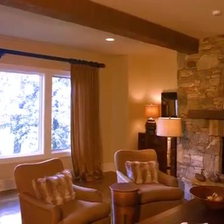}{0033} & \supprescene{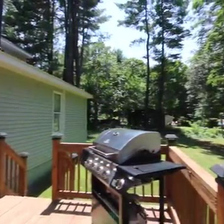}{0039} \\[1pt]
    \supprescene{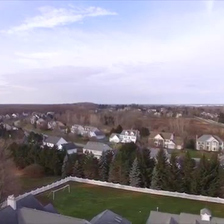}{0062} & \supprescene{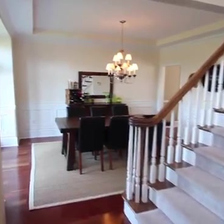}{0064} \\[1pt]
    \supprescene{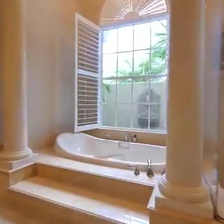}{0017} & \supprescene{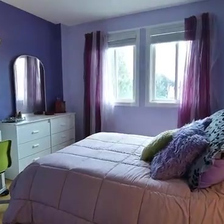}{0130} \\[1pt]
    \supprescene{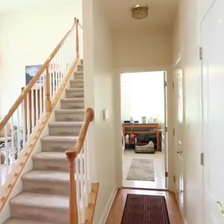}{0061} & \supprescene{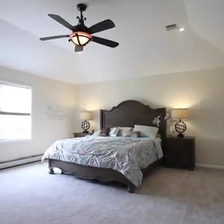}{0090} \\[1pt]
    \supprescene{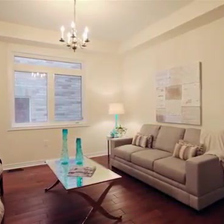}{0019} & \supprescene{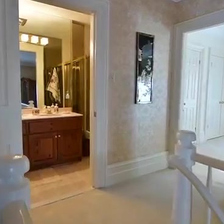}{0130} \\[1pt]
    \supprescene{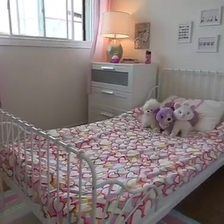}{0049} & \supprescene{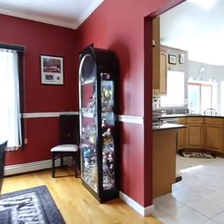}{0048} \\[1pt]
    \supprescene{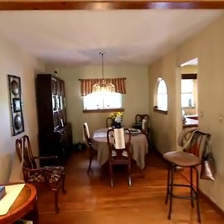}{0038} & \supprescene{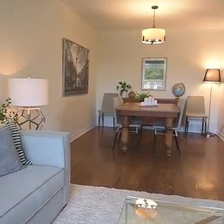}{0223} \\[1pt]
    \supprescene{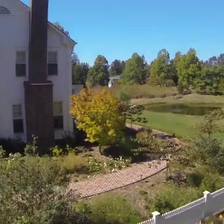}{0063} & \supprescene{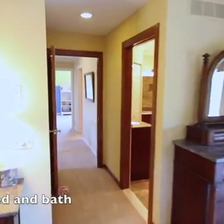}{0062} \\[1pt]
    \supprescene{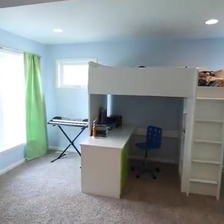}{0016} & \supprescene{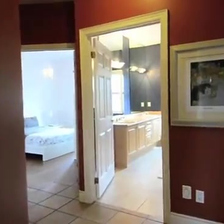}{0046}
  \end{tabular}
  }
  \caption{Random RealEstate10K visual samples, page 1. Each row shows two randomly selected scenes; each scene includes two input views and one target view, with columns ordered as LVSM, JiT-LVSM, and reference.}
  \label{fig:supp-random-re10k-a}
\end{figure}
\clearpage

\begin{figure}[H]
  \centering
  \setlength{\tabcolsep}{0.5pt}
  \renewcommand{\arraystretch}{0.2}
  \makebox[\textwidth][c]{%
  \begin{tabular}{c ccc @{\hspace{5pt}} c ccc}
    {\scriptsize Inputs} &
    {\scriptsize LVSM} & {\scriptsize JiT-LVSM} & {\scriptsize Reference} &
    {\scriptsize Inputs} &
    {\scriptsize LVSM} & {\scriptsize JiT-LVSM} & {\scriptsize Reference} \\
    \supprescene{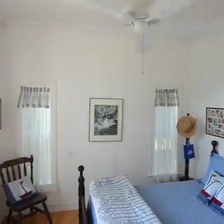}{0026} & \supprescene{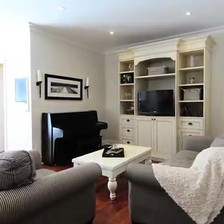}{0214} \\[1pt]
    \supprescene{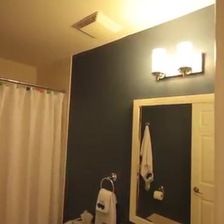}{0087} & \supprescene{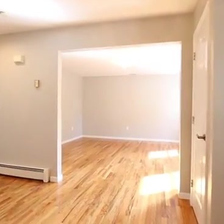}{0055} \\[1pt]
    \supprescene{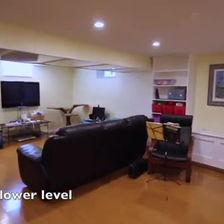}{0045} & \supprescene{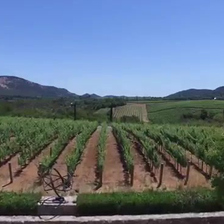}{0052} \\[1pt]
    \supprescene{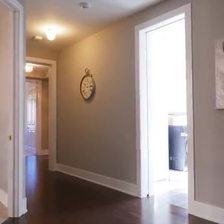}{0049} & \supprescene{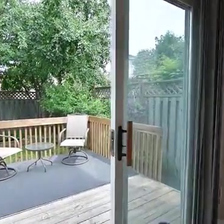}{0163} \\[1pt]
    \supprescene{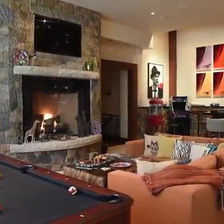}{0035} & \supprescene{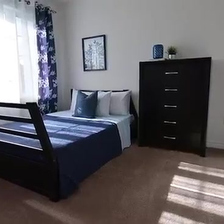}{0026} \\[1pt]
    \supprescene{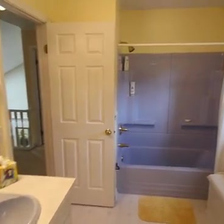}{0039} & \supprescene{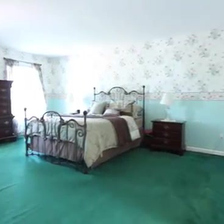}{0082} \\[1pt]
    \supprescene{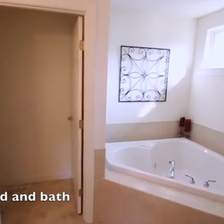}{0037} & \supprescene{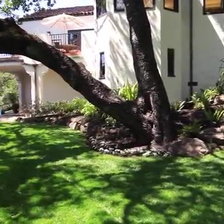}{0027} \\[1pt]
    \supprescene{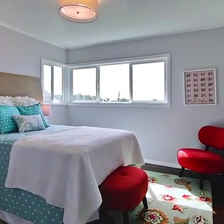}{0031} & \supprescene{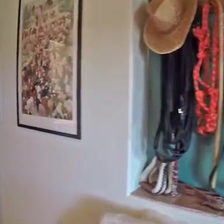}{0054} \\[1pt]
    \supprescene{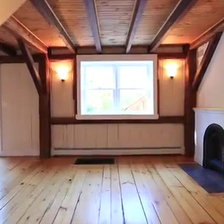}{0161} & \supprescene{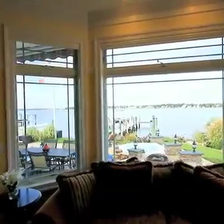}{0061} \\[1pt]
    \supprescene{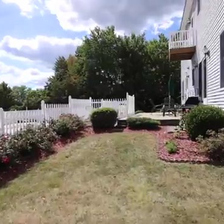}{0061} & \supprescene{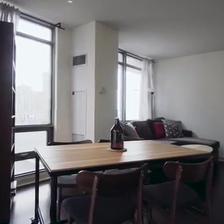}{0084}
  \end{tabular}
  }
  \caption{Random RealEstate10K visual samples, page 2. Each row shows two randomly selected scenes; each scene includes two input views and one target view, with columns ordered as LVSM, JiT-LVSM, and reference.}
  \label{fig:supp-random-re10k-b}
\end{figure}
\clearpage

\begin{figure}[H]
  \centering
  \setlength{\tabcolsep}{0.5pt}
  \renewcommand{\arraystretch}{0.2}
  \makebox[\textwidth][c]{%
  \begin{tabular}{c ccc @{\hspace{5pt}} c ccc}
    {\scriptsize Inputs} &
    {\scriptsize LVSM} & {\scriptsize JiT-LVSM} & {\scriptsize Reference} &
    {\scriptsize Inputs} &
    {\scriptsize LVSM} & {\scriptsize JiT-LVSM} & {\scriptsize Reference} \\
    \supprescene{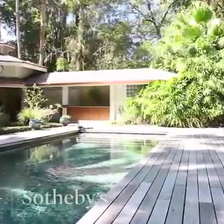}{0025} & \supprescene{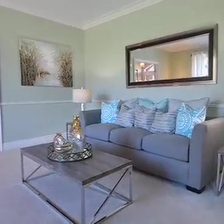}{0016} \\[1pt]
    \supprescene{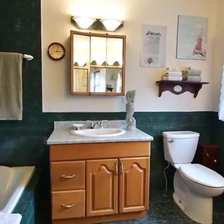}{0234} & \supprescene{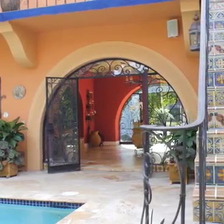}{0069} \\[1pt]
    \supprescene{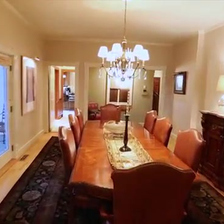}{0035} & \supprescene{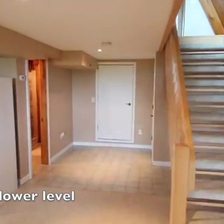}{0015} \\[1pt]
    \supprescene{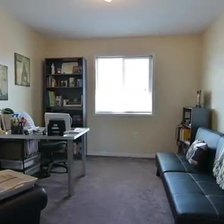}{0055} & \supprescene{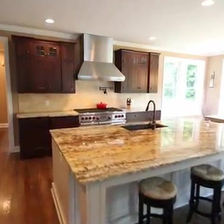}{0070} \\[1pt]
    \supprescene{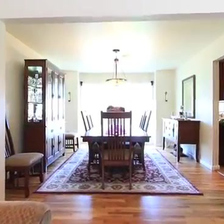}{0017} & \supprescene{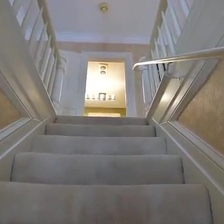}{0031} \\[1pt]
    \supprescene{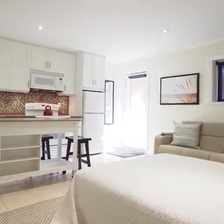}{0031} & \supprescene{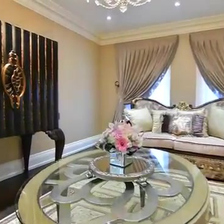}{0031} \\[1pt]
    \supprescene{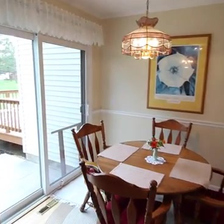}{0170} & \supprescene{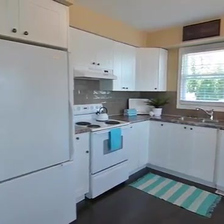}{0073} \\[1pt]
    \supprescene{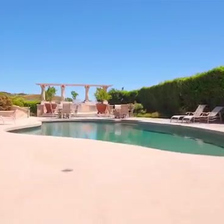}{0079} & \supprescene{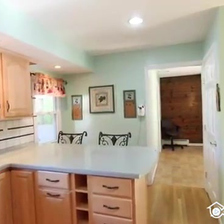}{0088} \\[1pt]
    \supprescene{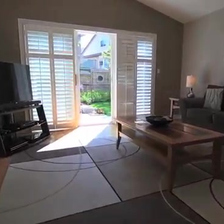}{0010} & \supprescene{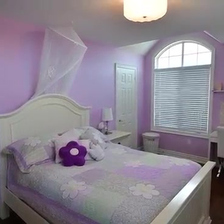}{0113} \\[1pt]
    \supprescene{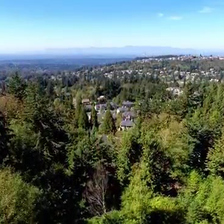}{0078} & \supprescene{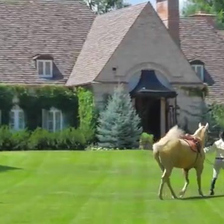}{0084}
  \end{tabular}
  }
  \caption{Random RealEstate10K visual samples, page 3. Each row shows two randomly selected scenes; each scene includes two input views and one target view, with columns ordered as LVSM, JiT-LVSM, and reference.}
  \label{fig:supp-random-re10k-c}
\end{figure}

\clearpage
\section{More Visual Results for JUSt3R}
\label{app:dust3r-gallery}

\begingroup
\setlength{\intextsep}{4pt}
\newcommand{\justpanel}[4]{%
  \vtop{\offinterlineskip\hsize=#1\relax
    \hbox to\hsize{\hfil{\scriptsize\strut #4}\hfil}%
    \kern1pt
    \vbox to#2{%
      \hbox to\hsize{\hfil\includegraphics[width=\hsize,height=#2,keepaspectratio]{#3}\hfil}%
      \vfil}}}
\newcommand{\justcase}[1]{%
  \vbox to149pt{\offinterlineskip
    \hbox to\textwidth{%
      \justpanel{0.160\textwidth}{44pt}{fig/dust3r/gallery/#1/rgb.png}{Input}\hfil
      \justpanel{0.160\textwidth}{44pt}{fig/dust3r/gallery/#1/gt.png}{Reference}\hfil
      \justpanel{0.160\textwidth}{44pt}{fig/dust3r/gallery/#1/ff.png}{DUSt3R}\hfil
      \justpanel{0.160\textwidth}{44pt}{fig/dust3r/gallery/#1/jit1.png}{JUSt3R$_1$}\hfil
      \justpanel{0.160\textwidth}{44pt}{fig/dust3r/gallery/#1/jit10.png}{JUSt3R$_{10}$}\hfil
      \justpanel{0.160\textwidth}{44pt}{fig/dust3r/gallery/#1/jit50.png}{JUSt3R$_{50}$}}%
    \kern11pt
    \justclouds{#1}{62pt}%
    \vfil}\par\nointerlineskip}
\newcommand{\justwidecase}[1]{%
  \vbox to149pt{\offinterlineskip
    \hbox to\textwidth{%
      \justpanel{0.326\textwidth}{30pt}{fig/dust3r/gallery/#1/rgb.png}{Input}\hfil
      \justpanel{0.326\textwidth}{30pt}{fig/dust3r/gallery/#1/gt.png}{Reference}\hfil
      \justpanel{0.326\textwidth}{30pt}{fig/dust3r/gallery/#1/ff.png}{DUSt3R}}%
    \kern5pt
    \hbox to\textwidth{%
      \justpanel{0.326\textwidth}{30pt}{fig/dust3r/gallery/#1/jit1.png}{JUSt3R$_1$}\hfil
      \justpanel{0.326\textwidth}{30pt}{fig/dust3r/gallery/#1/jit10.png}{JUSt3R$_{10}$}\hfil
      \justpanel{0.326\textwidth}{30pt}{fig/dust3r/gallery/#1/jit50.png}{JUSt3R$_{50}$}}%
    \kern11pt
    \justclouds{#1}{36pt}%
    \vfil}\par\nointerlineskip}
\newcommand{\justclouds}[2]{%
  \hbox to\textwidth{%
    \justpanel{0.244\textwidth}{#2}{fig/dust3r/gallery/#1/cloud_ff.jpg}{DUSt3R}\hfil
    \justpanel{0.244\textwidth}{#2}{fig/dust3r/gallery/#1/cloud_jit1.jpg}{JUSt3R$_1$}\hfil
    \justpanel{0.244\textwidth}{#2}{fig/dust3r/gallery/#1/cloud_jit10.jpg}{JUSt3R$_{10}$}\hfil
    \justpanel{0.244\textwidth}{#2}{fig/dust3r/gallery/#1/cloud_jit50.jpg}{JUSt3R$_{50}$}}}

\begin{figure}[H]
  \centering
  \justcase{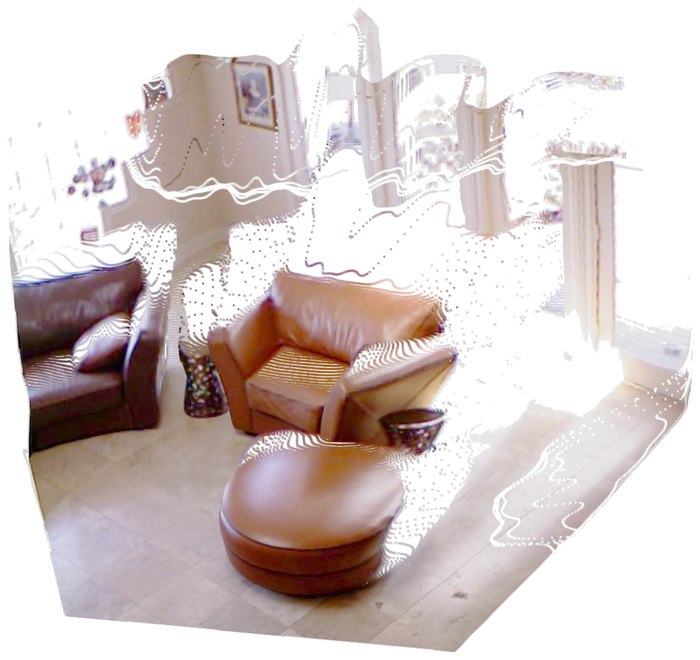}
  \justcase{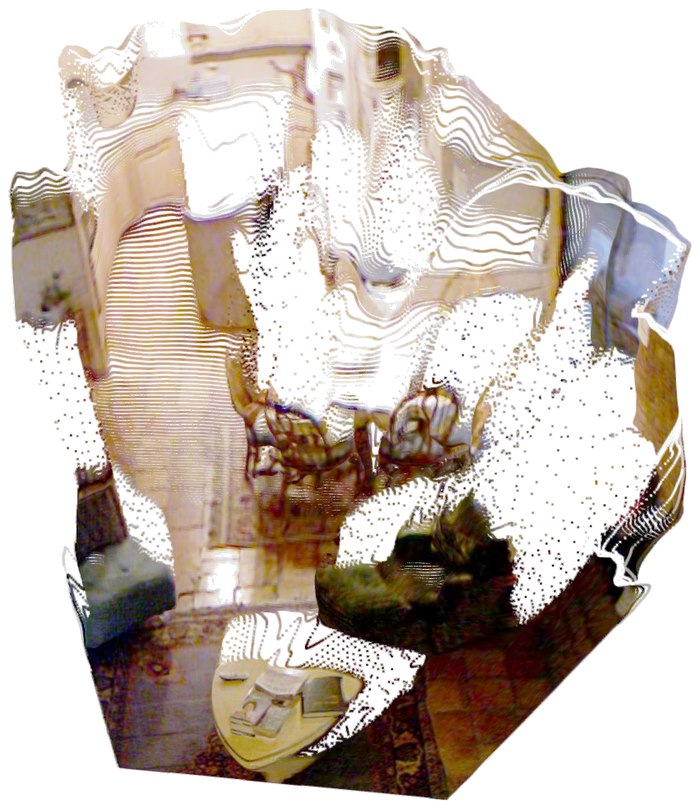}
  \justcase{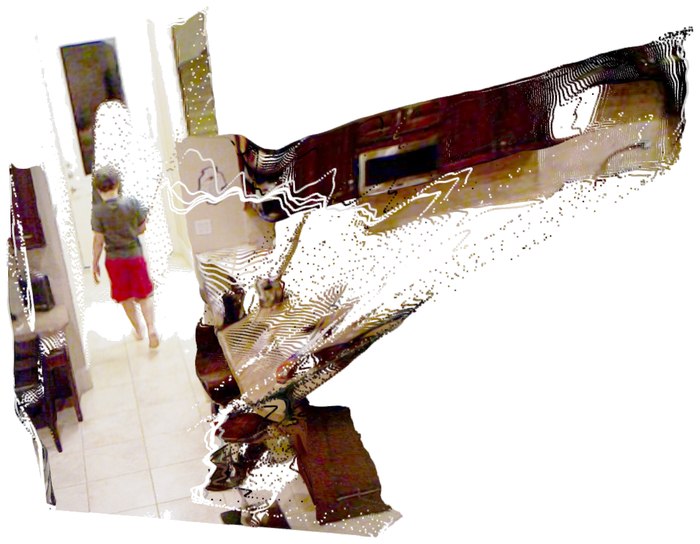}
  \justcase{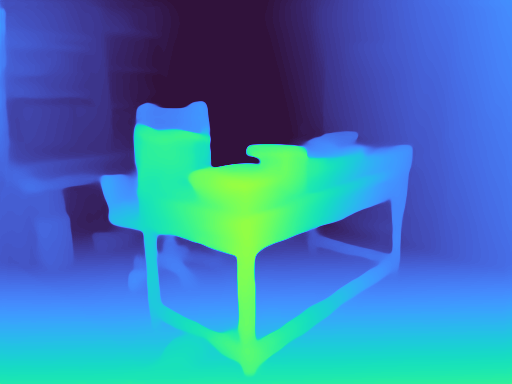}
  \caption{Additional JUSt3R comparisons, page 1. Each case shows depth predictions followed immediately by its offset point-cloud views.}
  \label{fig:just3r-gallery-page-1}
\end{figure}
\clearpage

\begin{figure}[H]
  \centering
  \justcase{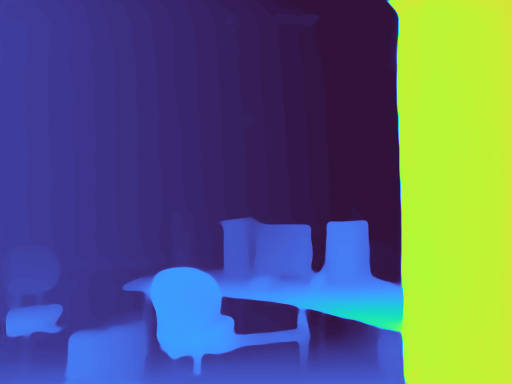}
  \justcase{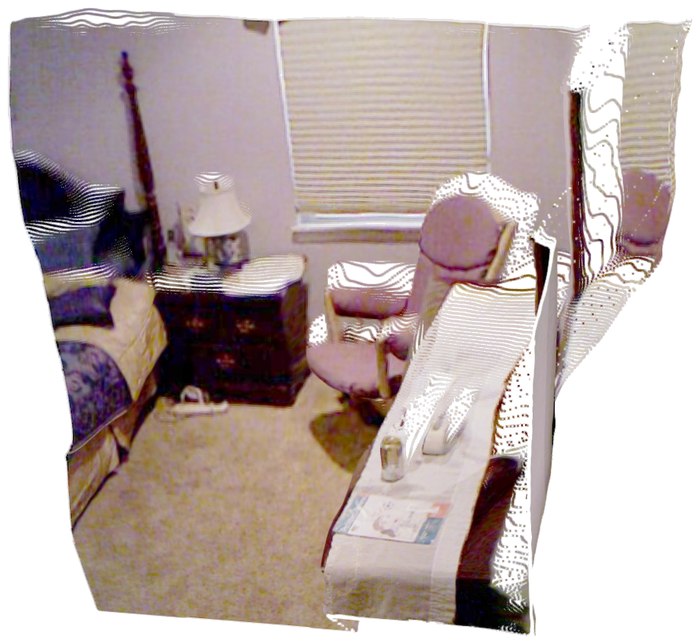}
  \justwidecase{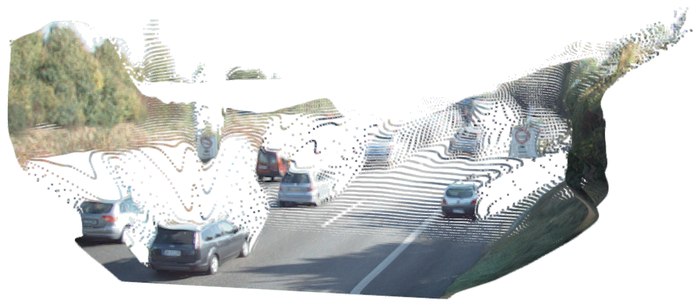}
  \justwidecase{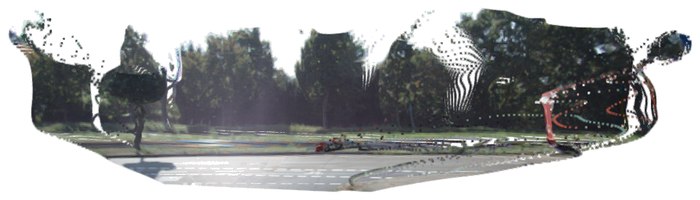}
  \caption{Additional JUSt3R comparisons, page 2. Each case shows depth predictions followed immediately by its offset point-cloud views.}
  \label{fig:just3r-gallery-page-2}
\end{figure}
\clearpage

\begin{figure}[H]
  \centering
  \justwidecase{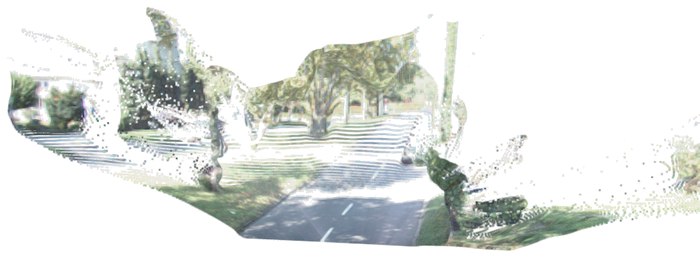}
  \justwidecase{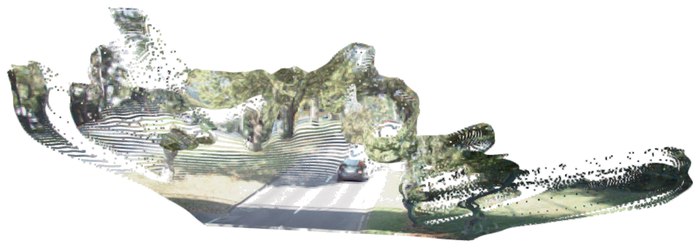}
  \justwidecase{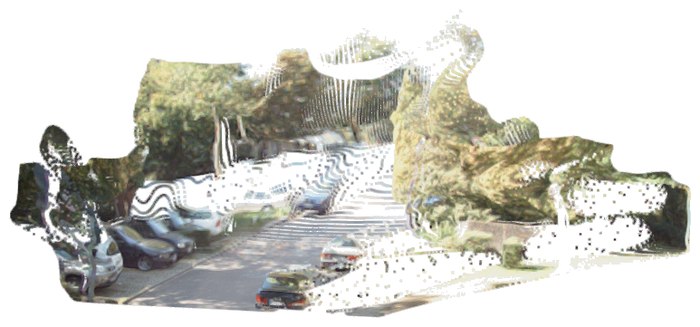}
  \justwidecase{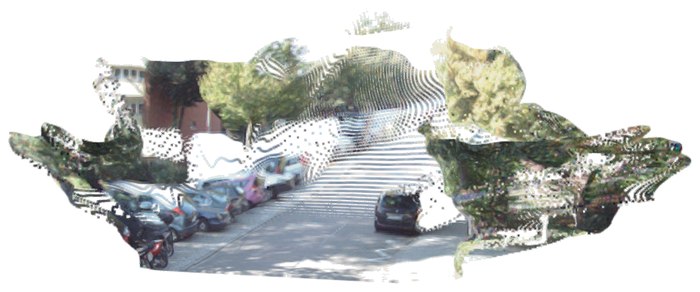}
  \caption{Additional JUSt3R comparisons, page 3. Each case shows depth predictions followed immediately by its offset point-cloud views.}
  \label{fig:just3r-gallery-page-3}
\end{figure}
\clearpage

\begin{figure}[H]
  \centering
  \justwidecase{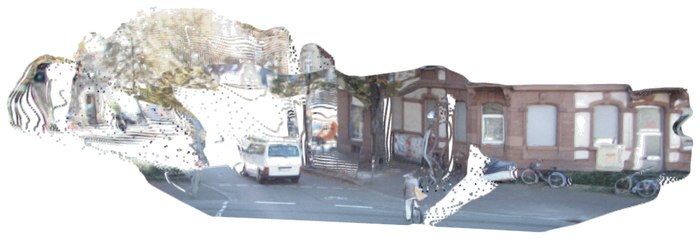}
  \justwidecase{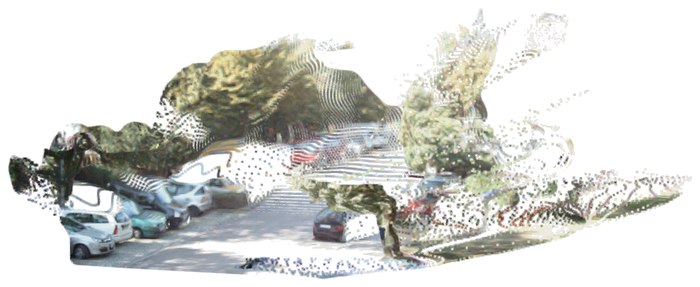}
  \justcase{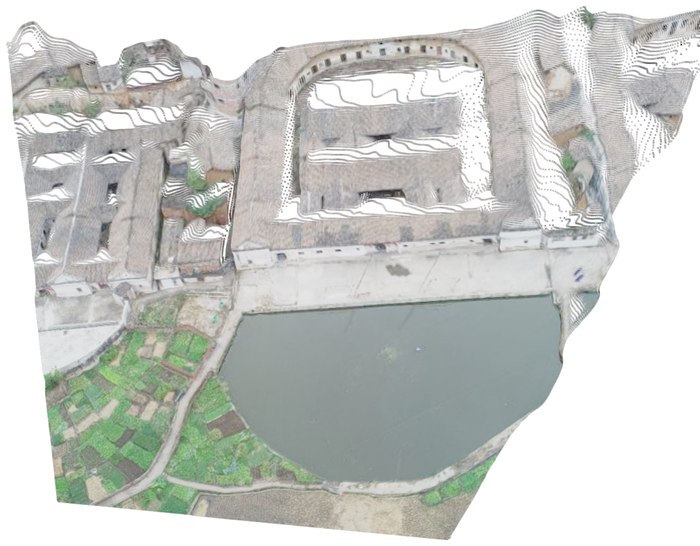}
  \justcase{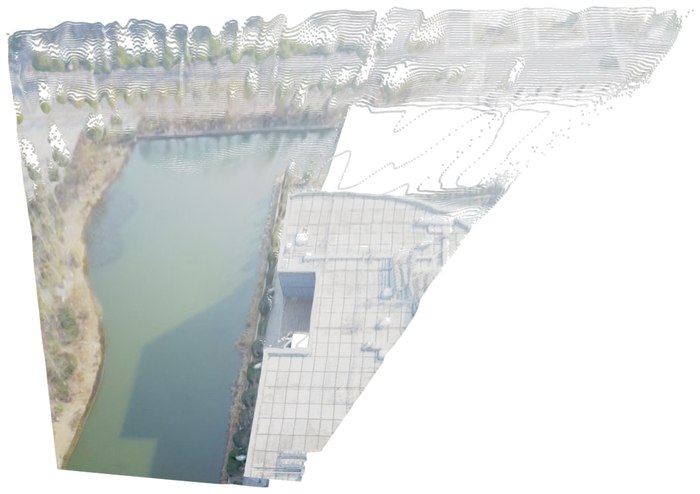}
  \caption{Additional JUSt3R comparisons, page 4. Each case shows depth predictions followed immediately by its offset point-cloud views.}
  \label{fig:just3r-gallery-page-4}
\end{figure}
\clearpage

\begin{figure}[H]
  \centering
  \justcase{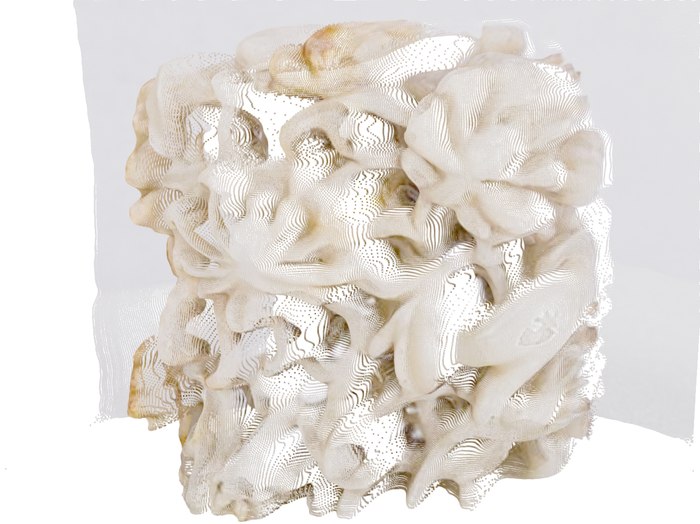}
  \justcase{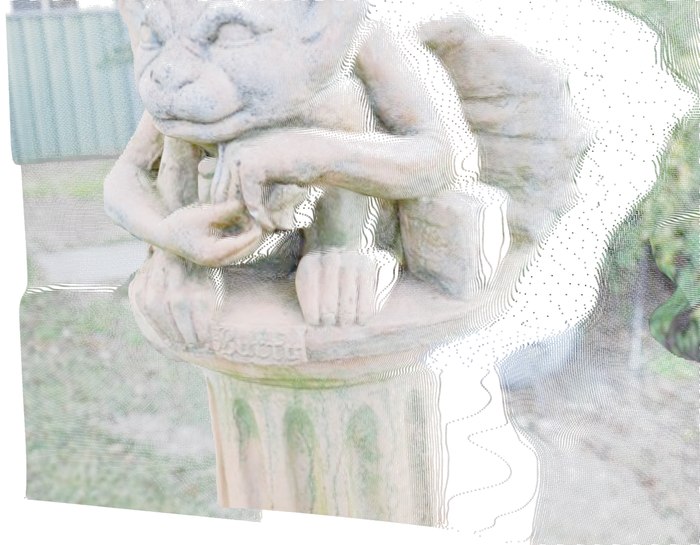}
  \justcase{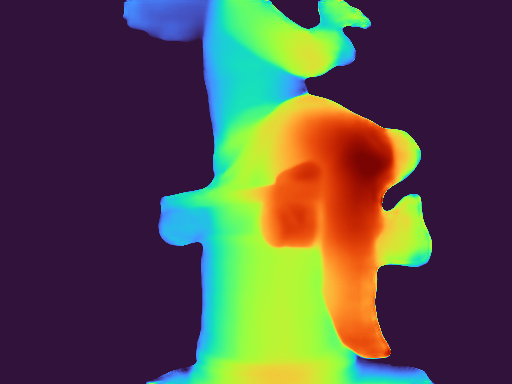}
  \caption{Additional JUSt3R comparisons, page 5. Each case shows depth predictions followed immediately by its offset point-cloud views.}
  \label{fig:just3r-gallery-page-5}
\end{figure}

\endgroup

\end{document}